\pdfoutput=1

\PassOptionsToPackage{table,xcdraw}{xcolor}

\documentclass[11pt]{article}

\usepackage[final]{acl}
\usepackage{xspace} 
\usepackage{times}
\usepackage{latexsym}
\usepackage[T1]{fontenc}
\usepackage[utf8]{inputenc}
\usepackage{longtable}
\usepackage{microtype}

\usepackage{hyperref}
\usepackage{booktabs}
\usepackage{graphicx}
\graphicspath{{./imgs/}}
\usepackage{footmisc}
\usepackage{microtype}

\usepackage{amsmath}
\usepackage{amsfonts,bm}
\usepackage{xspace}

\newcommand{\Ni}{({\em i})~}
\newcommand{\Nii}{({\em ii})~}
\newcommand{\Niii}{({\em iii})~}
\newcommand{\Niv}{({\em iv})~}
\newcommand{\Nv}{({\em v})~}

\definecolor{mypink3}{cmyk}{0, 0.7808, 0.4429, 0.1412}

\makeatletter   
\newcommand{\sveryshortarrow}[1][3pt]{\mathrel{%
    \vcenter{\hbox{\rule[-.5\fontdimen8\scriptfont3]
               {\scriptratio\dimexpr#1\relax}{\fontdimen8\scriptfont3}}}%
   \mkern-4mu\hbox{\let\f@size\sf@size\usefont{U}{lasy}{m}{n}\symbol{41}}}}
\makeatother

\def\eqref#1{equation~\ref{#1}}

\def\1{\bm{1}}

\def\m1{{\bm{1}}}

\DeclareMathAlphabet{\mathsfit}{\encodingdefault}{\sfdefault}{m}{sl}
\SetMathAlphabet{\mathsfit}{bold}{\encodingdefault}{\sfdefault}{bx}{n}

\def\gD{{\mathcal{D}}}

\usepackage{arydshln}
\usepackage{subcaption}
\usepackage{caption}
\usepackage{array, makecell} %
\usepackage{footmisc}
\usepackage{alltt}
\usepackage{floatrow}
\usepackage{pifont}
\usepackage{tabularx}
\usepackage{adjustbox}
\usepackage{multirow}

\usepackage{multirow, colortbl, caption}
\definecolor{lightblue}{RGB}{232, 244, 248}
\definecolor{lightpink}{RGB}{254, 238, 237}

\definecolor{bluelink}{RGB}{0,113,188}
\definecolor{greenlink}{RGB}{0,188,113}

\usepackage{tikz}

\usepackage{collcell}

\usepackage{etoolbox}

\newtoggle{inTableHeader}
\toggletrue{inTableHeader}
\newcommand*{\StartTableHeader}{\global\toggletrue{inTableHeader}}%

\let\OldTabular\tabular%
\let\OldEndTabular\endtabular%
\renewenvironment{tabular}{\StartTableHeader\OldTabular}{\OldEndTabular\StartTableHeader}%

\newcommand*{\MinNumber}{-1.0}%
\newcommand*{\MidNumber}{0.0} %
\newcommand*{\MaxNumber}{1.0}%

\newcommand{\ApplyGradient}[1]{%
  \iftoggle{inTableHeader}{#1}{
    \ifdim #1 pt > \MidNumber pt
        \pgfmathsetmacro{\PercentColor}{max(min(100.0*(#1 - \MidNumber)/(\MaxNumber-\MidNumber),100.0),0.00)} %
        \hspace{-0.33em}\colorbox{yellow!\PercentColor!blue}{#1}
    \else
        \pgfmathsetmacro{\PercentColor}{max(min(100.0*(\MidNumber - #1)/(\MidNumber-\MinNumber),100.0),0.00)} %
        \hspace{-0.33em}\colorbox{blue!\PercentColor!blue}{#1}
    \fi
  }}
\newcolumntype{R}{>{\collectcell\ApplyGradient}c<{\endcollectcell}}

\usepackage[nameinlink]{cleveref}
\crefformat{section}{\S#2#1#3} 
\crefname{algorithm}{Alg.}{Algs.}
\crefname{table}{Table}{Tables}
\crefformat{subsection}{\S#2#1#3}
\Crefname{equation}{Eq.}{Eqs.}
\Crefname{figure}{Figure}{Figures}

\usepackage[colorinlistoftodos,prependcaption,textsize=tiny]{todonotes}

\newcommand{\yes}{\textcolor{green!60!black}{\ding{51}}} 
\newcommand{\no}{\textcolor{red!60!black}{\ding{55}}}    
\newcommand{\partialsupported}{\textbf{\textcolor{orange}{$\sim$}}}

\usepackage{soul}

\usepackage{float}

\usepackage{multirow}
\usepackage{hhline}
\usepackage{amssymb}

\definecolor{headerLavender}{RGB}{230, 230, 250} 
\definecolor{rowLightGray}{RGB}{245, 245, 245} 
\definecolor{rowCream}{RGB}{255, 250, 240} 
\definecolor{errorRed}{RGB}{255, 77, 77} 

\definecolor{chartqapro1}{RGB}{30,160,220} 
\definecolor{chartqapro2}{RGB}{50,200,100} 
\title{
\raisebox{0.1em}{
\begin{adjustbox}{valign=c}
    \includegraphics[height=1.0em]{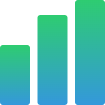}
\end{adjustbox}}
\hspace{-0.6em}
\textsc{\textbf{
\textcolor{chartqapro1}{ChartQA}\textcolor{chartqapro2}{Pro}
}}: A More Diverse and Challenging Benchmark \\ for Chart Question Answering
}

\newcommand{\chartqapro}[1]{\textsc{
\textcolor{chartqapro1}{ChartQA}\textcolor{chartqapro2}{Pro}}}

\author{
Ahmed Masry$^{\clubsuit}$ \thanks{\ \ Equal contribution.} \thanks{\ \ Corresponding author.}, Mohammed Saidul Islam$^{\clubsuit}$ \footnotemark[1], Mahir Ahmed$^{\clubsuit}$ \thanks{\ \ Equal contribution.}, Aayush Bajaj$^{\spadesuit}$\footnotemark[3] \\ \bf Firoz Kabir$^{\clubsuit}$\footnotemark[3], Aaryaman Kartha$^{\clubsuit}$\footnotemark[3], Md Tahmid Rahman Laskar$^{\clubsuit\heartsuit}$\footnotemark[3] \\ \bf Mizanur Rahman$^{\clubsuit\bigstar}$\footnotemark[3], 
Shadikur Rahman$^{\clubsuit}$\footnotemark[3], Mehrad Shahmohammadi$^{\clubsuit}$\footnotemark[3] \\ \bf Megh Thakkar$^{\spadesuit}$, Md Rizwan Parvez$^{\S}$, Enamul Hoque$^{\clubsuit}$, Shafiq Joty$^{\diamondsuit\triangle}$ \\
$^\clubsuit$York University, Canada, 
$^\heartsuit$Dialpad Inc., Canada, 
$^\bigstar$RBC, Canada  \\
$^\spadesuit$MILA - Quebec AI Institute, Canada, 
$^\S$Qatar Computing Research Institute (QCRI) \\
$^\diamondsuit$Nanyang Technological University, Singapore, 
$^\triangle$Salesforce Research, USA \\
\{masry20, saidulis, mrahmed, mdfkabir, aarykary\}@yorku.ca \\
\{tahmid20, mizanurr, shadikur, msm97, enamulh\}@yorku.ca \\
\{aayush.bajaj, megh.thakkar\}@mila.quebec, mparvez@hbku.edu.qa, sjoty@salesforce.com
}

\begin{document}
\maketitle

\begin{abstract} 
Charts are ubiquitous, as people often use them to analyze data, answer questions, and discover critical insights. However, performing complex analytical tasks with charts requires significant perceptual and cognitive effort. Chart Question Answering (CQA) systems automate this process by enabling models to interpret and reason with visual representations of data. However, existing benchmarks like ChartQA lack real-world diversity and have recently shown performance saturation with modern large vision-language models (LVLMs). To address these limitations, we introduce\chartqapro{}, a new benchmark that includes 1,341 charts from 157 diverse sources, spanning various chart types—including infographics and dashboards—and featuring 1,948 questions in various types, such as multiple-choice, conversational, hypothetical, and unanswerable questions, to better reflect real-world challenges. Our evaluations with 21 models show a substantial performance drop for LVLMs on\chartqapro{}; e.g., Claude Sonnet 3.5 scores \emph{90.5\%} on ChartQA but only \emph{55.81\%} on\chartqapro{}, underscoring the complexity of chart reasoning.  
We complement our findings with detailed error analyses and ablation studies, identifying key challenges and opportunities for advancing LVLMs in chart understanding and reasoning.
We release \chartqapro{} at \href{https://github.com/vis-nlp/ChartQAPro}{https://github.com/vis-nlp/ChartQAPro}.

\end{abstract}

\section{Introduction}

 \begin{figure}[t!]
    \includegraphics[width=\textwidth]{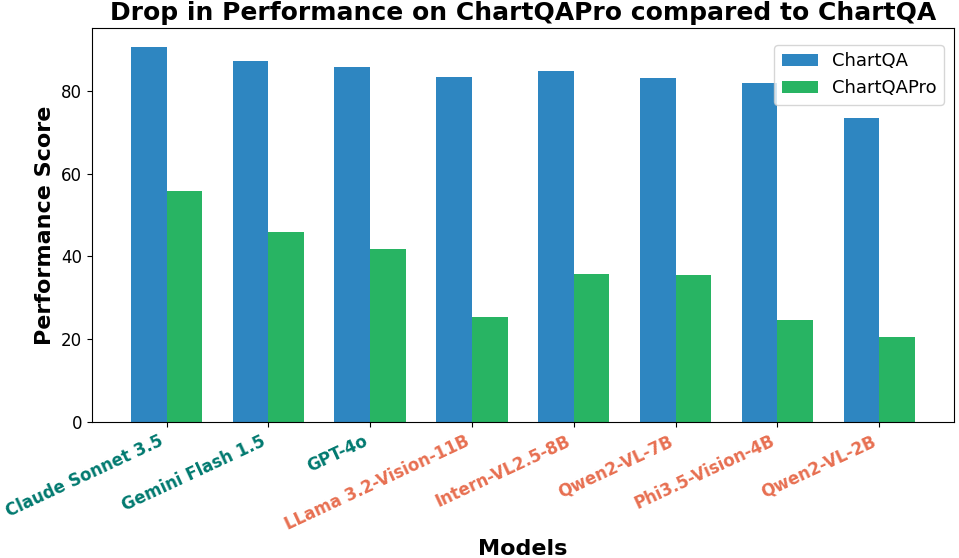}
    \caption{Performance gap between ChartQA \cite{masry-etal-2022-chartqa} and\chartqapro{} for various LVLMs. 
    }
    \label{fig:performance_gap}
\end{figure}

\begin{figure*}[t]
    \centering
    \includegraphics[width=\textwidth]{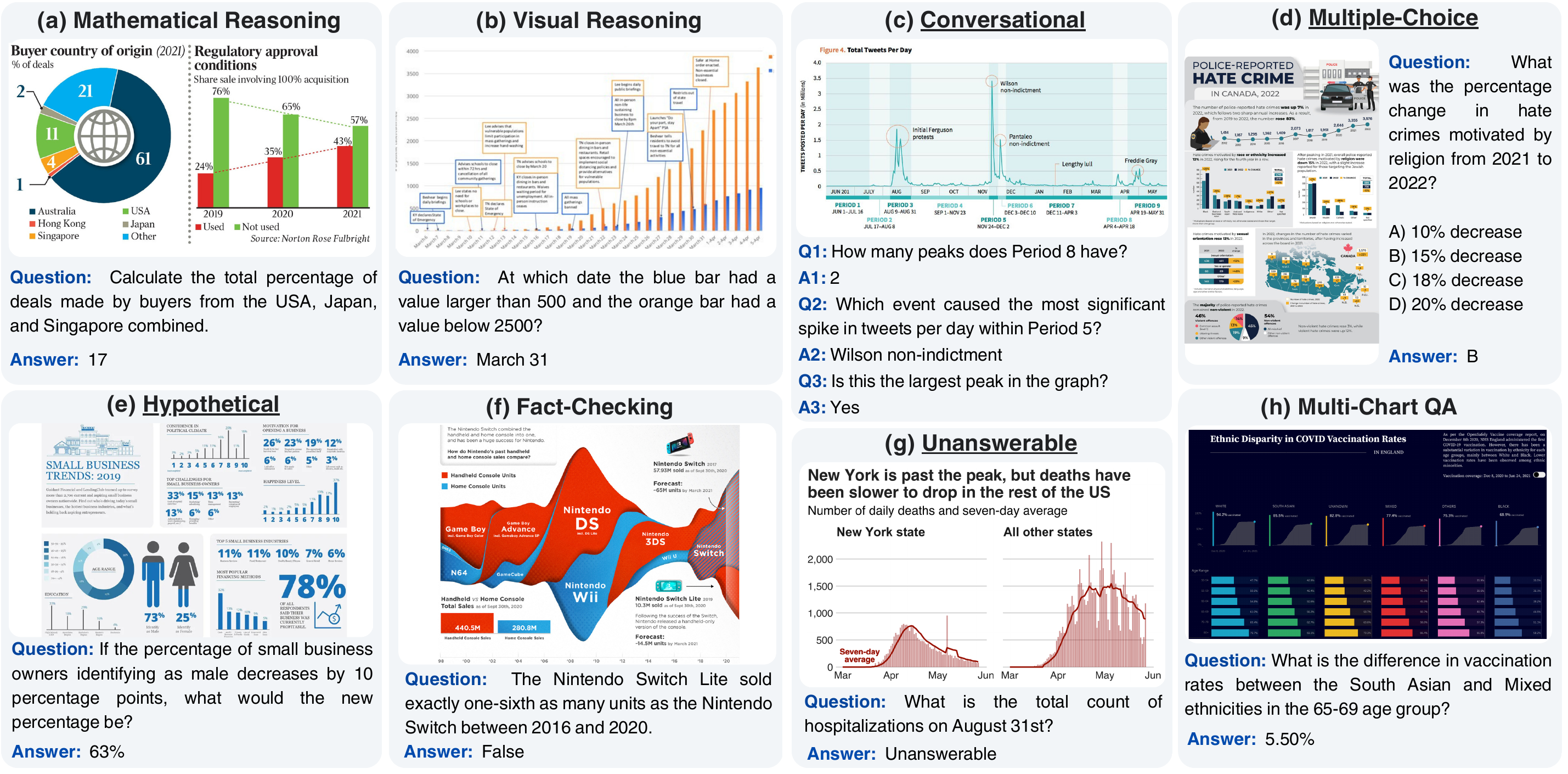}
    \caption{\label{fig:qa_types}\chartqapro{} covers a more diverse range of questions compared to existing chart question answering datasets (\cref{tab:datasets_comparison}), providing an extensive evaluation of chart understanding abilities. }
\end{figure*}

Data visualizations such as bar and line charts are very popular for analyzing data and making informed decisions across various domains such as finance, journalism, and science ~\cite{kim2020answering, masry2024chartgemma, hoque2022chartquestionansweringstate}. 
However, answering complex questions about charts can pose significant challenges as the user needs to combine visual perception with cognitive reasoning. Chart Question Answering (CQA) systems aim to assist users by taking questions about charts as input and generating answers. Unlike traditional visual question answering involving natural images and scenes, CQA requires models to interpret structured data visually, reason over relationships among visual elements and text, and derive contextual insights.

Due to its real-world relevance, CQA has become a key task for evaluating recent LVLMs~\cite{ 
Qwen2VL, openai2024gpt4technicalreport, geminiteam2024gemini15unlockingmultimodal, grattafiori2024llama3herdmodels}
. These LVLMs have obtained remarkable performance on multimodal tasks, including CQA. For instance, on ChartQA~\cite{masry-etal-2022-chartqa}, Claude Sonnet 3.5 \cite{claude} achieves an accuracy of 90.5\%, while GPT4~\cite{openai2024gpt4technicalreport} and Gemini~\cite{geminiteam2024gemini15unlockingmultimodal} 
reach 85.7\% and 87.2\%, respectively (Figure~\ref{fig:performance_gap}). Open-source LVLMs also  appear to be catching up, with Qwen2.5-VL~\cite{Qwen2VL} reporting 89.5\%. These striking results prompt two core questions: \Ni \emph{Is chart understanding and reasoning already a solved task?} and \Nii \emph{Have open-source models truly matched their closed-source counterparts?}

A closer look at ChartQA reveals key limitations. First, its chart images lack visual diversity, coming from a few online sources like Statista and Pew Research Center. It primarily includes only bar, line, and pie charts with numeric labels directly on visual elements, reducing the need for actual visual reasoning. Second, the benchmark focuses largely on factoid questions that require simple data extraction or basic arithmetic. Earlier datasets~\cite{figureqa, leafqa,stlcqa} suffer from similar issues, and are also curated from synthetic data or templated questions. Although a recent work, CharXiv~\cite{wang2024charxivchartinggapsrealistic}, addresses some of these limitations, it relies on charts sourced exclusively from papers on arXiv, limiting visual and topical diversity, and also lacking numerous real-world question types.

In contrast, real-world charts encompass diverse domains like economy, health, etc., and a wide variety of question types, including \emph{hypothetical} (e.g., future price prediction), \emph{multiple-choice} (e.g., in educational exams), \emph{conversational} (e.g., in decision-making meetings) and \emph{unanswerable} (e.g. due to missing data). Additionally, multi-chart layouts and dashboards are often used in finance, business intelligence, and scientific reports, requiring users to analyze multiple charts simultaneously. These types of questions and complex layouts are absent from current benchmarks, suggesting that existing evaluations do not fully capture the real-world challenges in chart understanding and create an overly optimistic perception of progress in this field.

To address these limitations and rigorously evaluate LVLMs’ on chart understanding, we present\chartqapro{}, a comprehensive benchmark of 1341 charts sourced from 157 diverse online platforms.\chartqapro{} includes 1948 human-written, human-verified question-answer pairs covering factoid, multiple-choice, conversational, hypothetical, multi-chart, and unanswerable queries, making it representative of real-world use cases  (see Figure \ref{fig:qa_types}). Beyond bar, line, and pie charts,\chartqapro{} features images with complex visualizations such as multi-chart layouts, infographics, and dashboards, introducing greater visual and analytical complexity. Inspired by conversational and multi-document QA in text such as CoQA~\cite{reddy2019coqaconversationalquestionanswering} and HotpotQA~\cite{yang2018hotpotqadatasetdiverseexplainable}, some questions also require multi-turn interactions or referencing accompanying paragraphs, probing a broader range of multimodal reasoning skills. 

\begin{table*}[t]
\centering
\small
\setlength{\tabcolsep}{4pt}

\definecolor{highlightrow}{RGB}{235,255,235} 

\caption{ 
Comparison of\chartqapro{} with existing chart-based QA benchmarks.
Features are grouped into \textbf{Chart Images} (real vs.\ synthetic data, number of sources, topic diversity, infographics/dashboards, accompanying paragraph, multi-chart support)
and \textbf{Questions Types} (MCQ, conversational, hypothetical, unanswerable).
\yes = Supported, \no = Not Supported, \partialsupported = Partially Supported. 
}
\label{tab:datasets_comparison}

\resizebox{\textwidth}{!}{%
\begin{tabular}{@{}l|cccccc|ccccc@{}}
\toprule
\rowcolor{gray!25}
 & \multicolumn{6}{c|}{\textbf{Chart Images}} & \multicolumn{5}{c}{\textbf{Question Types}} \\
\cmidrule(lr){2-7}\cmidrule(lr){8-12}
\textbf{Dataset} 
& \begin{tabular}[c]{@{}c@{}}\textbf{Real vs.}\\ \textbf{Synthetic}\end{tabular}
& \begin{tabular}[c]{@{}c@{}}\textbf{\# Chart}\\ \textbf{Sources}\end{tabular}
& \textbf{Topic Diversity}
& \begin{tabular}[c]{@{}c@{}}\textbf{Infographics}\\ \textbf{\& Dashboards}\end{tabular}
& \begin{tabular}[c]{@{}c@{}}\textbf{Accompanying}\\ \textbf{Paragraph}\end{tabular}
& \begin{tabular}[c]{@{}c@{}}\textbf{Multi}\\ \textbf{Chart}\end{tabular}
& \textbf{MCQ} 
& \textbf{Conversational} 
& \textbf{Hypothetical}
& \textbf{Unanswerable} 
& \textbf{Fact Checking} \\
\midrule
\rowcolor{gray!10}
\textbf{PlotQA} \cite{plotqa} 
& Synthetic 
& 1 
& \no 
& \no 
& \no 
& \no 
& \no 
& \no 
& \no 
& \no 
& \no \\[1ex]
\textbf{ChartQA} \cite{masry-etal-2022-chartqa}
& Real 
& 4 
& \partialsupported 
& \no 
& \no 
& \no 
& \no 
& \no 
& \no 
& \no 
& \no \\[1ex]
\rowcolor{gray!10}
\textbf{CharXiv} \cite{wang2024charxivchartinggapsrealistic}
& Real 
& 1 
& \no 
& \no 
& \no 
& \yes 
& \no 
& \no 
& \no 
& \yes 
& \no \\[1ex]
\rowcolor{highlightrow}
\textbf{\chartqapro{} (\textcolor{blue}{Ours})}
& Real 
& 157 
& \yes 
& \yes 
& \yes 
& \yes 
& \yes 
& \yes 
& \yes 
& \yes 
& \yes \\
\bottomrule
\end{tabular}%
}
\end{table*}

Our evaluations reveal a sharp performance drop for both closed- and open-source models on\chartqapro{} (Figure~\ref{fig:performance_gap}). For example, the SoTA Claude Sonnet 3.5’s accuracy falls from \emph{90.50\%} to \emph{55.81\%}, demonstrating that\chartqapro{} presents a more challenging and realistic benchmark for chart understanding, and that there is substantial room for improvement in LVLMs’ chart reasoning abilities. 
Moreover, while open-source models seemed to match closed-source ones on ChartQA, they still lag significantly on\chartqapro{}
with the best, Qwen2-VL-7B \cite{Qwen2VL}, achieving only \emph{37.17\%}. This suggests that prior benchmarks might have overstated progress due to their limited diversity.

Our contributions include:
\Ni a comprehensive benchmark that evaluates diverse and complex real-world chart understanding abilities;
\Nii extensive evaluation of open- and closed-source models, revealing significant performance declines compared to previous benchmarks;
\Niii in-depth qualitative analyses and ablation studies, identifying key challenges and future directions for improving LVLMs’ chart reasoning abilities.

\section{Related Work}

\paragraph{Chart Understanding Datasets} Numerous tasks and benchmarks have been developed to evaluate LVLMs’ chart understanding abilities, such as question answering~\cite{masry-etal-2022-chartqa, wang2024charxivchartinggapsrealistic}, chart summarization~\cite{kantharaj2022charttotextlargescalebenchmarkchart}, fact-checking~\cite{akhtar-etal-2023-reading, akhtar2023chartcheck}, and explanation generation~\cite{kantharaj2022opencqa}. Among these, chart question answering is the most commonly used for evaluation. Early benchmarks like STL-CQA~\cite{stlcqa} and Leaf-QA~\cite{leafqa} relied on synthetically generated charts and templated questions. Later benchmarks, such as ChartQA~\cite{masry-etal-2022-chartqa}, PlotQA~\cite{plotqa}, and CharXiv~\cite{wang2024charxivchartinggapsrealistic}, used real-world charts and more complex questions requiring advanced visual reasoning. However, these benchmarks extract charts from limited sources (\cref{tab:datasets_comparison}), cover few question types, and have reached performance saturation due to recent strong LVLMs (Figure~\ref{fig:performance_gap}). In contrast,\chartqapro{} sources from 157 diverse online domains and includes human-written, verified questions across multiple types (multiple-choice, conversational, hypothetical, etc.), offering a more challenging benchmark.

\paragraph{Vision-Language Models for Charts}
Advances in vision-language models have significantly improved chart understanding and reasoning. These models can be categorized into: \Ni closed-source, \Nii open-source general multimodal models, and \Niii chart-specific models.  Closed-source models~\cite{openai2024gpt4technicalreport, geminiteam2024gemini15unlockingmultimodal} achieve the highest performance on recent chart understanding benchmarks ~\cite{masry-etal-2022-chartqa, wang2024charxivchartinggapsrealistic}. Open-source general multimodal models ~\cite{Qwen2VL, li2024llava, internvl, wu2024deepseekvl2mixtureofexpertsvisionlanguagemodels, abdin2024phi3technicalreporthighly, idefics3, masry2025alignvlmbridgingvisionlanguage, rodriguez2024bigdocsopenpermissivelylicenseddataset} currently lag behind, but are rapidly closing the gap. Chart-specific models ~\cite{masry2024chartgemma, masry2024chartinstruct, zhang2024tinychartefficientchartunderstanding, masry2023unichartuniversalvisionlanguagepretrained} demonstrate strong performance on standard benchmarks ~\cite{masry-etal-2022-chartqa, akhtar2023chartcheck, kantharaj2022charttotextlargescalebenchmarkchart, ahmed-workshop-2021}. However, their generalization to real-world chart understanding remains uncertain due to their reliance on instruction-tuning datasets with limited task diversity. 
\chartqapro{} offers a more comprehensive benchmark, ensuring that model improvements reflect real progress in chart understanding abilities of these models.

\section{\textsc{The\chartqapro{} Benchmark}}
\begin{figure*}[t]
    \centering
    \includegraphics[width=\textwidth]{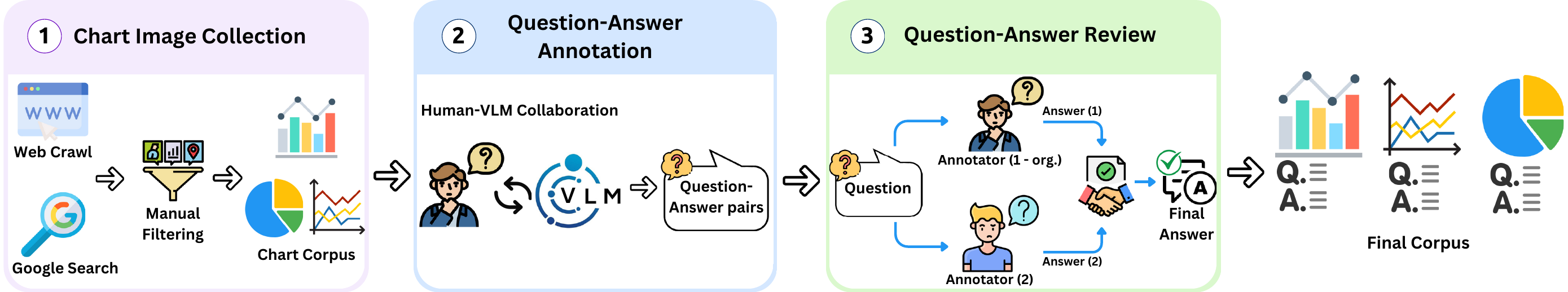}
    \caption{\chartqapro{} Dataset Construction Process 
    }
    \label{fig:data_const}
\end{figure*}

\subsection{Dataset Construction}
Our dataset construction pipeline consists of three key stages (see \Cref{fig:data_const}): \Ni Chart Image Collection, \Nii Question-Answer Annotation, and \Niii Question-Answer Review. We detail each stage below:

\paragraph{Stage 1 - Chart Images Collection}
\chartqapro{} prioritizes both visual and topical diversity. We sourced chart images from diverse platforms featuring real-world visualizations, including multi-series line charts, stacked and grouped bar charts, dashboards, and infographics. Key sources include Pew Research \cite{pewresearch}, Tableau \cite{tableaupublic}, the Public Policy Institute of California (PPIC) \cite{ppic}, and Our World in Data (OWID) \cite{owid} (see \Cref{fig:src_charts} for more details). 
For Pew and Tableau, we randomly sampled charts from \citet{islam-etal-2024-datanarrative} which are already diverse in visual styles, while for other sources, we manually selected charts with varied formats to enhance dataset diversity. Some charts were accompanied by textual descriptions that provided additional context, improving the interpretability of the corresponding chart images.

To further expand coverage, we collected an additional 1041 charts from the web, building upon prior efforts from ChartInstruct \cite{masry2024chartinstruct} to include dashboards and infographics. In total,\chartqapro{} is a compiled dataset of 1341 chart images from \textbf{157 online platforms}, covering a broad spectrum of chart types and styles. Additional details are provided in \Cref{app:AMT}.

\paragraph{Stage 2 - Question-Answer Annotation}

\chartqapro{} includes five types of question-answer pairs: \Ni Reasoning, \Nii Conversational, \Niii Multiple-Choice, \Niv Hypothetical, and \Nv Fact-Checking. Nine team members collaboratively created these QA pairs, with five focusing on reasoning questions and the remaining four handling other categories. To ensure high-quality annotations, we adopted a human-VLM collaboration process for each QA type:

\begin{itemize}
    \item \textbf{Curating Seed QA pairs:} Annotators crafted a diverse set of seed QA pairs covering different question types that required complex reasoning.
    \item \textbf{VLM-Assisted Expansion:} Using GPT-4o, Gemini, and Claude, we expanded the seed set by generating additional QA pairs. We decided to employ multiple models to mitigate bias. Each model was prompted with a seed QA pair and tasked with generating five new pairs per chart. In addition, annotators interactively prompted VLMs to generate additional QA pairs beyond those derived from the seed set, encouraging the models to produce diverse and novel questions.
    \item \textbf{Human Refinement:} Annotators manually reviewed the generated questions to filter the ones that are overly simple (e.g., direct data retrieval from charts) or revise the questions that are unclear or ambiguous. 
\end{itemize}

A key feature of\chartqapro{} is the inclusion of \textit{unanswerable questions}. These questions were carefully curated by humans to be closely related to the chart’s topic while unanswerable based solely on the chart image. Also,\chartqapro{} features questions on chart-text pairs, with some referring only to the chart, others only to the text, and some requiring integration of both, posing a greater challenge for vision-language models. 
We present a brief description of various question types below:\\
\noindent \textbf{Reasoning:} 
Reasoning with charts is a common real-world task involving visual perception, trend analysis, and mathematical reasoning. While such questions appear in benchmarks like ChartQA, we focus on more complex cases requiring compositional calculations and deeper pattern, trend, and outlier analysis (e.g., Figure \ref{fig:qa_types}a, b).

\noindent \textbf{Conversational:}
Conversational questions consist of multiple interrelated QA pairs for a given visualization, where each question naturally builds upon the previous one. These questions help us assess how well VLMs handle contextual dependencies, such as coreference resolution and logical or arithmetic reasoning (e.g., Figure \ref{fig:qa_types}c). 

\noindent \textbf{Multiple-Choice:} Multiple-choice questions (MCQs) are widely used in assessments and educational materials. We focused on MCQs that require complex reasoning, including trend analysis, anomaly detection, extrapolation, and time series analysis (e.g., Figure \ref{fig:qa_types}d). 

Each question is presented with four answer choices, covering various formats such as dates, percentages, locations, and specific labels derived from the data.

\noindent \textbf{Hypothetical:} Hypothetical questions introduce assumptions beyond observable chart data (e.g., Figure \ref{fig:qa_types}e). Answering these questions requires not only extracting information accurately but also making inferences, estimations, or approximations based on patterns and trends present in the visualization. These questions add an extra layer of complexity by requiring the model to reason beyond explicit data points.

\noindent \textbf{Fact-Checking:} Fact-checking questions involve evaluating a claim about a chart by extracting and verifying relevant data (e.g., Figure \ref{fig:qa_types}f). Each claim is classified as either \textit{True} (confirmed by data) or \textit{False} (contradicted by data). These questions test the model’s ability to interpret chart information and assess the validity of claims, a crucial skill for misinformation detection, incorrect prediction, fake news detection, etc.

\paragraph{Stage 3 - Question-Answer Review} 
After creating the QA pairs, we conducted a quality assessment to ensure accuracy and clarity.  Seven annotators, all co-authors with expertise in visualization, performed this review. Five focused on factoid questions, while the remaining two handled other categories.
Each annotator reviewed questions from a category they had not originally worked on, then cross-checked their responses with the category’s original creator. Any identified errors in the questions or answers were collaboratively revised until both parties reached an agreement. In rare instances, ambiguous questions were modified to resolve disagreements. For subjective questions (e.g., value estimations), minor discrepancies (<1\%) were considered acceptable. Overall, the initial agreement rate between annotators was 66.17\% before resolving all discrepancies.

\subsection{Dataset Analysis}
\subsubsection{Visual Diversity}
\begin{figure}[t]
    \centering
    \includegraphics[width=\columnwidth]{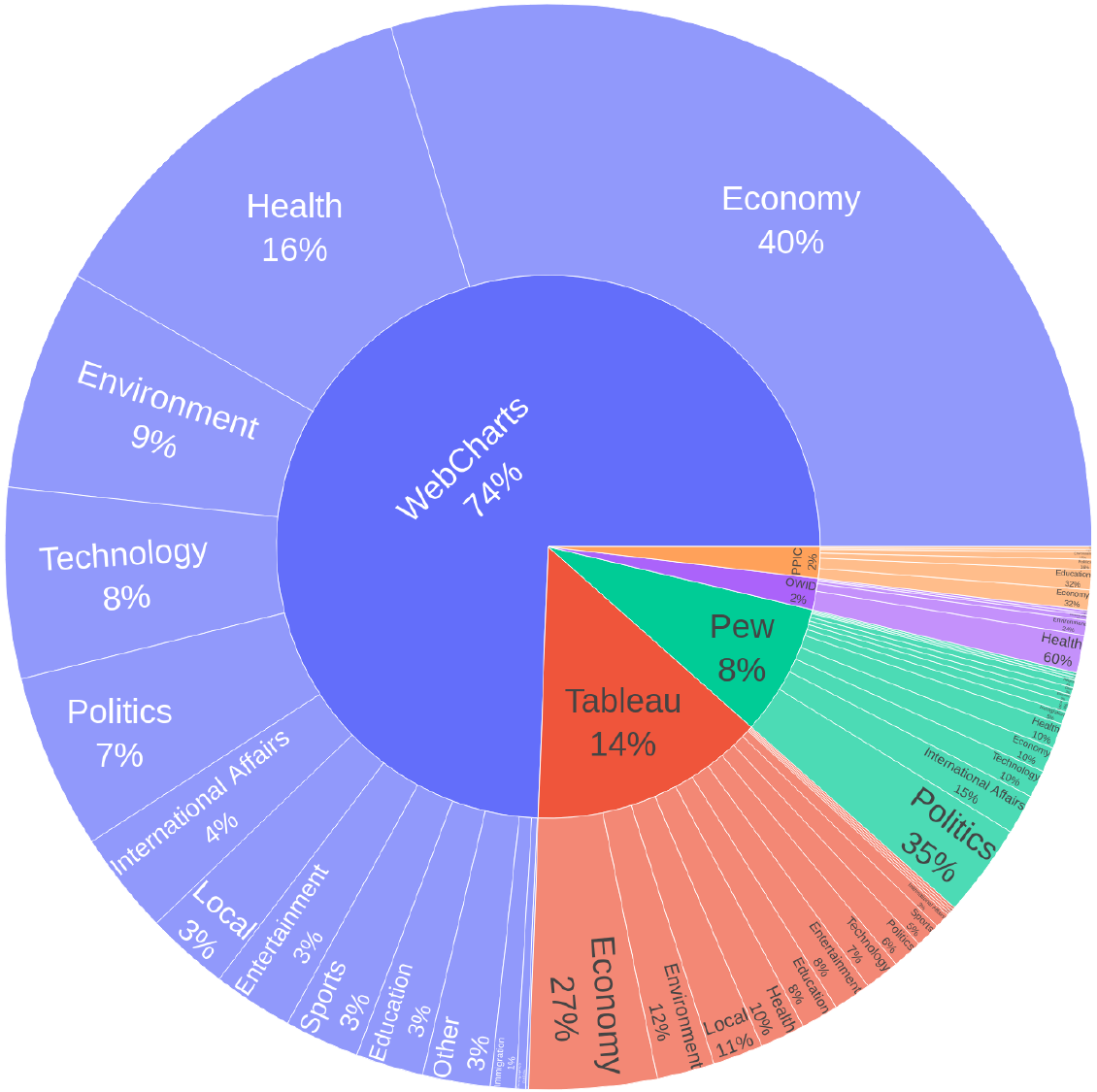}
     \caption{Distribution of topics per source in\chartqapro{}. The inner ring represents online sources, while the outer ring shows topic distribution for each source.
    }
    \label{fig:ch_src}
\end{figure}

Unlike the ChartQA \cite{masry-etal-2022-chartqa} dataset, which sources its charts from only four origins, our benchmark incorporates a diverse range of sources. These include web charts collected from various websites and links across the internet, as well as charts from Tableau, Pew Research, PPIC, and OWID. As shown in \Cref{fig:ch_src}, the majority of charts (74\%) were collected through web crawling, followed by charts from Tableau (14\%), 
covering a diverse range of topics, such as,
`Politics', `Economy', `Health', `Environment', `Technology', etc. The corpus also includes various
chart types such as bars, lines, pies, scatter plots, dashboards, infographics, maps, etc. (see \Cref{tab:qtype_ctype}), with bar charts being the most common (31.8\%), followed by line charts (26.5\%). 

To further quantify the visual diversity of our chart images compared to earlier benchmarks—ChartQA \cite{masry-etal-2022-chartqa} and CharXiv \cite{wang2024charxivchartinggapsrealistic}—we conducted an experiment where we first encoded all images from each benchmark into feature vectors using a CLIP vision encoder \cite{clip} with sentence-transformers \cite{reimers-2019-sentence-bert}. For each benchmark, we then computed the \emph{pairwise cosine distances} among all images. In this context, a higher average pairwise distance indicates that the images are less similar and therefore more visually diverse. Our\chartqapro{} benchmark exhibits an average distance of \emph{0.53}, while ChartQA and CharXiv show averages of \emph{0.26} and \emph{0.27}, respectively. Moreover, Figure \ref{fig:visual_diversity} in \ref{app:data_analysis} shows that most pairwise distances in\chartqapro{} exceed those in the other benchmarks. These results conclusively demonstrate that our\chartqapro{} benchmark is significantly more diverse than the existing benchmarks, offering a richer and more varied set of visual representations.

\begin{table*}[]
\centering
\caption{Distribution of chart and question types in\chartqapro{}.
}
\label{tab:qtype_ctype}
\resizebox{\columnwidth}{!}{%
\begin{tabular}{@{}lccccccccc|ccccc@{}}\rowcolor{gray!25} 
                                       & \multicolumn{9}{c}{\textbf{Chart Types}}
                                       & \multicolumn{5}{c}{\textbf{Question Types}} \\ \midrule 
  &
  \textbf{Bar} &
  \textbf{Line} &
  \textbf{Pie} &
  \textbf{Area} &
  \textbf{Scatter} &
  \textbf{Bubble} &
  \textbf{Dashboard} &
  \textbf{Infographic} &
  \textbf{Other} &
  \begin{tabular}[c]{@{}c@{}}\textbf{Math} \textbf{\&} \textbf{Visual}\\ \textbf{Reasoning}\end{tabular} &
  \textbf{Conversational} &
  \begin{tabular}[c]{@{}c@{}}\textbf{Fact}\\ \textbf{Checking}\end{tabular} &
  \begin{tabular}[c]{@{}c@{}}\textbf{Multiple}\\ \textbf{Choice}\end{tabular} &
  \multicolumn{1}{c}{\textbf{Hypothetical}} \\ \midrule
  \rowcolor{gray!10}
\multicolumn{1}{l}{\textbf{Count}}  & 427 & 355 & 29 & 30 & 8 & 7 & 258 & 190 & 37 & 1081 & 311 & 244 & 214 & \multicolumn{1}{c}{98} \\ \Xhline{1pt}
\end{tabular}%
}
\end{table*}

\subsubsection{Linguistic Diversity}
We conducted a detailed analysis of the linguistic features of our benchmark dataset (see Appendix \ref{app:data_analysis}). 
Unlike existing chart-based benchmarks that focus on short question-answer pairs,\chartqapro{} provides a more diverse and linguistically rich dataset. It features 6,638 unique tokens in questions and 1496 in answers, significantly surpassing CharXiv (4545) and ChartQA (2427). The questions in\chartqapro{} are longer and more varied, averaging 106.05 characters and 18 tokens, compared to CharXiv (96.3 characters, and 17.2 tokens) and ChartQA (63.25 characters, and 11.5 tokens), while answers remain concise at 6.7 characters and 1.18 tokens. Additionally,\chartqapro{} captures real-world variability with diverse syntactic structures, informal language, and typographical errors, making it a comprehensive benchmark for evaluating complex question-answering models in the chart domain.

We further analyze the linguistic diversity and richness of the text in chart images by extracting text using the Google OCR API\footnote{\url{https://cloud.google.com/vision/docs/ocr}} and using two key metrics: \textbf{lexical diversity} and \textbf{semantic diversity} (Figure \ref{fig:linguistic_diversity}). Lexical diversity, measured via the type-token ratio (TTR), is highest for\chartqapro{} (\emph{0.15}), followed by ChartQA (\emph{0.13}) and ChartXiv (\emph{0.11}), indicating a richer vocabulary in\chartqapro{}. Semantic diversity, quantified as the average pairwise cosine distance between text embeddings computed using sentence transformers~\cite{reimers-2019-sentence-bert}, is also maximum for\chartqapro{} (\emph{0.84}) compared to ChartQA (\emph{0.75}) and ChartXiv (\emph{0.78}), suggesting broader semantic coverage. 
Overall, these findings collectively demonstrate that\chartqapro{} exhibits greater linguistic diversity than previous benchmarks. More details are provided in \ref{app:visual_diversity}.

\section{Experiments}
\subsection{Problem Formulation}
\label{sec:problem-extraction}
We formulate the\chartqapro{} tasks as multimodal question-answering challenges. The dataset consists of $N$ examples, denoted as $\gD = \{c_i, q_i, a_i\}_{i=1}^N$, where each example includes a chart image $c_i$, a question $q_i$, and the corresponding ground truth answer $a_i$. For certain charts, the formulation also includes a corresponding context paragraph $p_i$ which the task might use. The objective is for the multimodal LLM to take $c_i$ and $q_i$ as input (along with the prompt) and autoregressively generate the answer $a_i$. We provide all our prompts in \ref{app:prompts-evaluation} to ensure reproducibility and transparency. 

\definecolor{human_baseline}{RGB}{255, 245, 170}
\definecolor{open_models_below_4B}{RGB}{185, 235, 255}
\definecolor{open_models_7B_12B}{RGB}{255, 219, 187}
\definecolor{closed_models}{RGB}{240, 240, 240}
\definecolor{chart_specific_models}{RGB}{217, 240, 211}

\begin{table*}[t]
\centering
\caption{Accuracy (\%) on\chartqapro{} by Prompt Type (main headers) and Question Type (sub-headers). Each Prompt Type block has five question types plus an Overall sub-column. Color coding for comparison: \colorbox{human_baseline!70}{human baseline}, \colorbox{closed_models!70}{closed-source models}, \colorbox{open_models_below_4B!50}{open-source models below 7B parameters}, \colorbox{open_models_7B_12B!50}{open-source models between 7-12B parameters}, \colorbox{chart_specific_models!70}{chart-specific models}. We bold the best score within each model category. }

\resizebox{\textwidth}{!}{%
\begin{tabular}{l|cccccc|cccccc|cccccc}
\toprule
\multirow{2}{*}{\textbf{Model}} 
& \multicolumn{6}{c|}{\textbf{Direct}} 
& \multicolumn{6}{c|}{\textbf{Chain-of-Thought (CoT)}} 
& \multicolumn{6}{c}{\textbf{Program-of-Thought (PoT)}} 
\\
\cmidrule(lr){2-7} \cmidrule(lr){8-13} \cmidrule(lr){14-19}
& \textbf{Factoid} & \textbf{MCQ} & \textbf{Convers.} & \textbf{FactChk.} & \textbf{Hypoth.} & \textbf{Overall}
& \textbf{Factoid} & \textbf{MCQ} & \textbf{Convers.} & \textbf{FactChk.} & \textbf{Hypoth.} & \textbf{Overall}
& \textbf{Factoid} & \textbf{MCQ} & \textbf{Convers.} & \textbf{FactChk.} & \textbf{Hypoth.} & \textbf{Overall} \\
\midrule\rowcolor{human_baseline!70}
Human Baseline          & 80.00 & 94.00 & 88.70 & 92.00 & 70.42 & \cellcolor{human_baseline}85.02 & N/A & N/A & N/A & N/A & N/A & N/A & N/A & N/A & N/A & N/A & N/A & N/A \\
\midrule
\multicolumn{19}{l}{\textbf{\textit{Closed-Source Models}}} \\

\rowcolor{closed_models!50} GPT4-o                   & 35.76 & 46.72 & 34.75 & 45.49 & 28.91 & \cellcolor{closed_models}37.67 & 37.40 & 61.68 & 33.93 & 57.37 & 30.83 & \cellcolor{closed_models}41.68 & 39.22 & 42.99 & 38.62 & 44.67 & 44.43 & \cellcolor{closed_models}40.48 \\

\rowcolor{closed_models!50}Gemini-Flash-2.0          & \textbf{43.43} & \textbf{60.28} & 40.25 & \textbf{67.62} & 24.47 & \cellcolor{closed_models}\textbf{46.85} & 51.51 & 69.15 & \textbf{43.84} & \textbf{67.62} & 39.89 & \cellcolor{closed_models}53.66 & \textbf{51.18} & \textbf{57.00} & \textbf{46.34} & \textbf{56.81} & 44.86 & \cellcolor{closed_models}\textbf{51.44} \\

\rowcolor{closed_models!50} Gemini-Flash-1.5        & 39.96 & 57.00 & 39.70 & 47.13 & 45.31 & \cellcolor{closed_models}42.96 & 42.37 & 64.01 & 40.17 & 56.14 & 39.42 & \cellcolor{closed_models}45.97 & 45.57 & 35.51 & 40.98 & 50.40 & \textbf{47.26} & \cellcolor{closed_models}44.42 \\

\rowcolor{closed_models!50} Claude Sonnet 3.5       & 38.84 & 51.40 & \textbf{44.53} & 55.60 & \textbf{45.48} & \cellcolor{closed_models}43.58 & \textbf{53.61} & \textbf{78.03} & \textbf{43.84} & 65.16 & \textbf{46.11} & \cellcolor{closed_models}\textbf{55.81} & 46.58 & 54.20 & 46.17 & 52.04 & 46.90 & \cellcolor{closed_models}48.05 \\
\midrule
\multicolumn{19}{l}{\textbf{\textit{Open-Source Models}}} \\

\rowcolor{open_models_below_4B!50} Intern-VL2.5-1B  & 9.15 & 7.00 & 6.20 & 16.63 & 8.17 & \cellcolor{open_models_below_4B}9.33 & 5.45 & 0.46 & 14.86 & 21.17 & 17.08 & \cellcolor{open_models_below_4B}8.96 & 1.07 & 0.0 & 0.64 & 0.40 & 2.04 & \cellcolor{open_models_below_4B}0.85 \\

\rowcolor{open_models_below_4B!50} Janus-1.3B             & 4.56 & 1.86 & 6.74 & 40.98 & 5.31 & \cellcolor{open_models_below_4B}9.21 & 3.54 & 0.0 & 6.05 & 29.91 & 6.97 & \cellcolor{open_models_below_4B}7.03 & 5.12 & 1.86 & 6.61 & 3.68 & 3.60 & \cellcolor{open_models_below_4B}4.74 \\

\rowcolor{open_models_below_4B!50} Qwen-VL2-2B             & 15.90 & 27.57 & 24.26 & 34.42 & 12.82 & \cellcolor{open_models_below_4B}20.68 & \textbf{16.62} & 30.84 & 23.89 & \textbf{38.52} & 13.00 & \cellcolor{open_models_below_4B}\textbf{21.90} & 13.66 & 23.83 & 15.22 & 8.60 & 3.06 & \cellcolor{open_models_below_4B}13.86 \\
\rowcolor{open_models_below_4B!50} Intern-VL2.5-2B & 13.86 & 10.74 & 14.02 & 45.90 & 18.92 & \cellcolor{open_models_below_4B}17.81 & 9.42 & 6.07 & 13.02 & 36.06 & 19.23 & \cellcolor{open_models_below_4B}13.46 & 1.13 & 6.07 & 2.51 & 2.04 & 3.06 & \cellcolor{open_models_below_4B}2.10 \\
\rowcolor{open_models_below_4B!50} SmolVLM-2.3B  & 13.32 & 16.82 & 17.71 & \textbf{46.31} & 25.21 & \cellcolor{open_models_below_4B}19.14 & 13.03 & 7.47 & 18.60 & 36.88 & 22.15 & \cellcolor{open_models_below_4B}16.76 & 4.03 & 12.61 & 11.22 & 5.73 & 12.52 & \cellcolor{open_models_below_4B}6.76 \\
\rowcolor{open_models_below_4B!50} Ovis1.6-LLama3.2-3B     & 12.87 & 0.46 & 4.18 & 40.98 & 10.17 & \cellcolor{open_models_below_4B}13.50 & 14.43 & 7.45 & 8.37 & 35.27 & 16.60 & \cellcolor{open_models_below_4B}15.42 & \textbf{17.41} & 5.60 & 5.86 & \textbf{30.32} & \textbf{24.10} & \cellcolor{open_models_below_4B}\textbf{16.22} \\

\rowcolor{open_models_below_4B!50} DeepSeek-VL2-3.4B          & 12.20 & 7.47 & 19.40 & 36.88 & 19.21 & \cellcolor{open_models_below_4B}16.28 & 9.63 & 1.40 & 18.09 & 38.11 & \textbf{23.25} & \cellcolor{open_models_below_4B}14.33 & 10.27 & 3.27 & 15.94 & 22.54 & 17.43 & \cellcolor{open_models_below_4B}12.30 \\

\rowcolor{open_models_below_4B!50} Phi 3.5-Vision-4B       & \textbf{17.48} & \textbf{30.37} & \textbf{28.54} & 41.99 & \textbf{37.27} & \cellcolor{open_models_below_4B}\textbf{24.73} & 10.55 & \textbf{32.71} & \textbf{27.20} & 8.19 & 8.16 & \cellcolor{open_models_below_4B}15.23 & 10.34 & \textbf{32.71} & \textbf{16.62} & 0.0 & 5.10 & \cellcolor{open_models_below_4B}12.24 \\

\rowcolor{open_models_7B_12B!50} Qwen-VL2-7B             & 30.70 & \textbf{44.85} & \textbf{35.68} & 48.36 & \textbf{37.23} & \cellcolor{open_models_7B_12B}35.59 & \textbf{32.95} & 46.26 & \textbf{37.60} & 50.40 & 29.65 & \cellcolor{open_models_7B_12B}\textbf{37.17} & 11.74 & \textbf{44.85} & \textbf{20.42} & 28.96 & 10.64 & \cellcolor{open_models_7B_12B}18.86 \\
\rowcolor{open_models_7B_12B!50} Intern-VL2.5-8B             & \textbf{35.21} & 25.70 & 32.26 & \textbf{53.27} & 29.61 & \cellcolor{open_models_7B_12B}\textbf{35.67} & 29.53 & 23.36 & 28.87 & \textbf{56.14} & 27.73 & \cellcolor{open_models_7B_12B}31.99 & \textbf{26.14} & 18.69 & 11.43 & \textbf{34.83} & 22.60 & \cellcolor{open_models_7B_12B}\textbf{23.88} \\
\rowcolor{open_models_7B_12B!50} Idefics-3-LLama-3.1-8B  & 20.69 & 2.29 & 31.96 & 10.76 & 36.83 & \cellcolor{open_models_7B_12B}20.03 & 20.06 & 2.29 & 30.98 & 11.14 & \textbf{35.36} & \cellcolor{open_models_7B_12B}19.51 & 10.06 & 5.41 & 19.41 & 7.62 & 18.60 & \cellcolor{open_models_7B_12B}11.16 \\
\rowcolor{open_models_7B_12B!50} LLaVA-Next-Mistral-7B   & 15.35 & 35.98 & 21.09 & 41.80 & 17.79 & \cellcolor{open_models_7B_12B}21.97 & 9.43 & 4.20 & 19.30 & 38.93 & 21.71 & \cellcolor{open_models_7B_12B}14.74 & 4.93 & 2.33 & 3.72 & 13.79 & 13.26 & \cellcolor{open_models_7B_12B}5.98 \\

\rowcolor{open_models_7B_12B!50} Ovis1.6-Gemma2-9B      & 30.25 & 4.67 & 28.93 & 27.86 & 28.21 & \cellcolor{open_models_7B_12B}26.83 & 18.09 & 12.42 & 17.68 & 25.05 & 20.49 & \cellcolor{open_models_7B_12B}18.39 & 22.59 & 20.56 & 17.33 & 32.37 & \textbf{25.30} & \cellcolor{open_models_7B_12B}22.89 \\
\rowcolor{open_models_7B_12B!50} LLama 3.2-Vision-11B           & 12.34 & 2.33 & 0.19 & 27.18 & 10.93 & \cellcolor{open_models_7B_12B}11.09 & 19.65 & \textbf{47.66} & 19.15 & 44.45 & 13.10 & \cellcolor{open_models_7B_12B}25.43 & 19.69 & 39.25 & 19.28 & 27.45 & 23.72 & \cellcolor{open_models_7B_12B}22.95 \\

\midrule
\multicolumn{19}{l}{\textbf{\textit{Chart-Specific Models}}} \\
\rowcolor{chart_specific_models!70} ChartGemma-3B              & 6.86 & 0.0 & 16.00 & 1.22 & 6.53 & \cellcolor{chart_specific_models}6.84 & \textbf{11.01} & 1.86 & \textbf{15.21} & 2.45 & \textbf{15.02} & \cellcolor{chart_specific_models}9.80 & \textbf{12.69} & 0.0 & \textbf{10.14} & \textbf{14.18} & \textbf{21.61} & \cellcolor{chart_specific_models}\textbf{11.52} \\
\rowcolor{chart_specific_models!70} TinyChart-3B               & \textbf{8.52} & \textbf{7.00} & \textbf{17.46} & \textbf{33.19} & \textbf{16.06} & \cellcolor{chart_specific_models}\textbf{13.25} & 8.97& \textbf{6.07} & 11.05 & \textbf{28.27} & 14.24 & \cellcolor{chart_specific_models}\textbf{11.67} & 5.64 & 0.0 & 4.11 & 0.0 & 15.92 & \cellcolor{chart_specific_models}4.59 \\
\rowcolor{chart_specific_models!70} ChartInstruct-LLama2-7B    & 7.09 & 0.0 & 3.77 & 0.0 & 6.91 & \cellcolor{chart_specific_models}4.88 & 3.83 & 0.0 & 4.43 & 0.40 & 10.65 & \cellcolor{chart_specific_models}3.42 & 0.09 & \textbf{0.31} & 1.69 & 2.04 & 0.0 & \cellcolor{chart_specific_models}0.61 \\
\bottomrule
\end{tabular}
}
\label{tab:prompt-question-accuracy}
\end{table*}

\subsection{Models} \label{subsec:data2text}
To evaluate the current state-of-the-art in chart understanding, we benchmark a diverse set of closed- and open-source models. The closed-source models include: \emph{(i)} GPT-4o~\cite{openai2024gpt4technicalreport}, \emph{(ii)} Gemini-Flash-1.5 and 2.0 ~\cite{geminiteam2024gemini15unlockingmultimodal}, and \emph{(iii)} Claude Sonnet 3.5~\cite{claude}. For open-source models, we categorize them based on parameter size. Models with fewer than 7B parameters include: \emph{(i)} Intern-VL2.5-1B \cite{internvl}, \emph{(ii)} Janus-1.3B \cite{wu2024janusdecouplingvisualencoding} \emph{(iii)} Qwen-VL2-2B \cite{Qwen2VL}, \emph{(iv)} Intern-VL2.5-2B \cite{internvl}, \emph{(v)} SmolVLM-2.3B \cite{smolvlm}, \emph{(vi)} Ovis1.6-Llama3.2-3B \cite{ovis}, \emph{(vii)} DeepSeek-VL2-3.4B \cite{wu2024deepseekvl2mixtureofexpertsvisionlanguagemodels}, and \emph{(viii)} Phi 3.5-Vision-4B \cite{abdin2024phi3technicalreporthighly}. In the 7-12B parameter range, we evaluate: \emph{(i)} Qwen-VL2-7B \cite{Qwen2VL}, \emph{(ii)} Intern-VL2.5-8B \cite{internvl}, \emph{(iii)} Idefics-3-Llama-3.1-8B \cite{idefics3}, \emph{(iv)} LLaVA-Next-Mistral-7B \cite{li2024llava}, \emph{(v)} Ovis1.6-Gemma2-9B\cite{ovis}, and \emph{(vi)} Llama 3.2-Vision-11B \cite{grattafiori2024llama3herdmodels}. In addition, we also evaluate chart-specific LVLMs: (i) ChartGemma ~\cite{masry2024chartgemma}, (ii) ChartInstruct-LLama2 ~\cite{masry2024chartinstruct}, (iii) TinyChart ~\cite{zhang2024tinychartefficientchartunderstanding}. All models are assessed with three prompting strategies: Direct prompting, Chain-of-Thought (CoT) \cite{wei2023chainofthoughtpromptingelicitsreasoning}, and Program-of-Thought (PoT) \cite{chen2023programthoughtspromptingdisentangling}. All experiments were run on Google Cloud Platform (GCP) using A100 GPU.

\subsection{Evaluation Metric} \label{subsec:evalmetric}
We enhance the relaxed accuracy metric commonly used for CQA~\cite{masry-etal-2022-chartqa, plotqa} for all the question types. Specifically, for numeric answers, we maintain a 5\% error margin, but for answers in `years' we require an exact match to avoid bias from minimal differences (e.g., 2008 vs. 2009). For textual answers (e.g., labels or common words), we employ the ANLS score \cite{biten2019scenetextvisualquestion}. Finally, multiple-choice questions (e.g., a, b, c, d) and fact-checking tasks (true, false) are evaluated using an exact-match criterion. Additional details are provided in ~\ref{app:eval-metric}.

\subsection{Main Results}
Table \ref{tab:prompt-question-accuracy} presents each model’s performance on the\chartqapro{} dataset under three prompting strategies (Direct, Chain-of-Thought, and Program-of-Thought) and across five question types. Closed‐source models consistently outperform open‐source counterparts in all prompting setups, and they also benefit from more extensive reasoning strategies (CoT or PoT), which boost overall accuracy. Notably, Chain‐of‐Thought yields the highest scores, with Claude Sonnet 3.5 achieving the top accuracy of \emph{55.81\%}, while GPT4o ranks lowest among the closed‐source group. We also observe that conversational, hypothetical, and factoid queries pose the greatest challenge for these models, whereas fact‐checking and multiple‐choice questions yield relatively higher accuracy—likely because the narrower range of possible answers increases the likelihood of a correct response.

In contrast, open‐source models below 7B parameters (highlighted in blue) exhibit substantially lower performance across all prompt types, often falling below 20\% overall accuracy. However, certain open‐source models in the 7–12B range (shaded in orange) show more promise; for instance, Qwen2‐VL‐7B and InternVL‐2.5‐8B both exceed 30\%. Surprisingly, these models often perform worse when asked to produce long‐form reasoning (as in CoT or PoT), suggesting they may lack sufficient training or alignment with step‐by‐step answer styles. Finally, chart‐specific models perform poorly under all setups, indicating that they may be heavily overfitted to particular visual and question types and thus generalize poorly to broader chart‐based QA scenarios.

Overall, these findings indicate that none of the models have achieved near‐human‐level chart understanding (See \ref{app:human_baseline}), leaving considerable room for improvement—a result that contrasts sharply with the previously reported high accuracies on previous datasets (Figure \ref{fig:performance_gap} and Appendix \ref{app:comparison_with_prev_benchs}).

\begin{figure*}[t]
    \includegraphics[width=\textwidth]{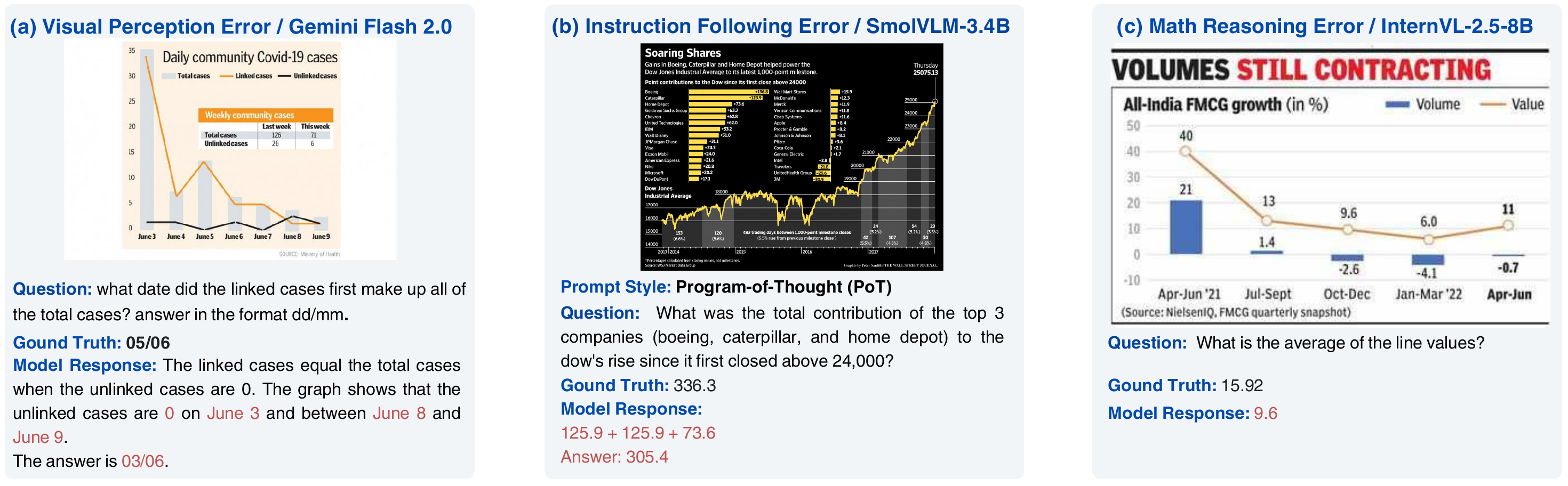}
    \caption{Sample errors across three categories: Visual Perception, Instruction Following, and Math Reasoning.}
    \label{fig:three_error_samples}
\end{figure*}

\subsection{Qualitative Analysis}
We examined 150 random samples to find common failure patterns and discovered three major error categories. Figure~\ref{fig:three_error_samples} presents representative errors, while additional examples are provided in ~\ref{app:error_analysis}.

\noindent \textbf{Visual Perception: } A common source of error is the failure to accurately recognize data values from chart images. This often occurs when charts are overcrowded with visual elements (e.g., bars, lines) or when data values are not explicitly shown, requiring inference based on geometric properties like height or area. While both open-source and closed-source models struggle with visual perception, it is the primary issue for closed-source models.

\noindent \textbf{Instruction Following: } Open-source and chart-specific models struggle to generate proper chain-of-thought (CoT) or program-of-thought (PoT) responses when explicitly prompted. Many generated programs even fail to execute due to runtime errors. Additionally, Llama 3.2 Vision-11B~\cite{grattafiori2024llama3herdmodels} performs poorly in the direct-answer setup (11.09\% accuracy), often ignoring the prompt and persistently generating CoT explanations, suggesting overfitting to CoT-style training.

\noindent \textbf{Math Reasoning: } While all models struggle with complex mathematical operations in our benchmark, closed-source models mitigate this issue to some extent by effectively utilizing long reasoning traces, such as Chain-of-Thought (CoT) or Program-of-Thought (PoT), allowing them to break down problems into steps and leverage external tools (e.g., Python). In contrast, open-source models fail to utilize these prompting strategies. In the direct-answer setup, they particularly struggle to perform multiple mathematical operations and generate the final answer correctly.

\begin{table}[t!]
\centering

\caption{
  Ablation results on\chartqapro{} across three independent dimensions.
  \textcolor{black}{\colorbox[HTML]{DAE8FC}{\textbf{(A) Chart Type}}}, \textcolor{black}{\colorbox[HTML]{FFF2CC}{\textbf{(B) Answer Type}}}, \textcolor{black}{\colorbox[HTML]{D5E8D4}{\textbf{(C) Paragraph Presence}}}.}
\label{tab:ablation-3-in-1}

\resizebox{\textwidth}{!}{
\begin{tabular}{
    l|
    >{\columncolor[HTML]{DAE8FC}}c
    >{\columncolor[HTML]{DAE8FC}}c
    >{\columncolor[HTML]{DAE8FC}}c|
    >{\columncolor[HTML]{FFF2CC}}c
    >{\columncolor[HTML]{FFF2CC}}c|
    >{\columncolor[HTML]{D5E8D4}}c
    >{\columncolor[HTML]{D5E8D4}}c
}
\toprule
 & \multicolumn{3}{c|}{\textbf{Chart Type (A)}} 
 & \multicolumn{2}{c|}{\textbf{Answer Type (B)}}
 & \multicolumn{2}{c}{\textbf{Paragraph Presence (C)}} \\
\cmidrule(lr){2-4}\cmidrule(lr){5-6}\cmidrule(lr){7-8}
\textbf{Model} & Chart & Dashboard & Infographic 
               & Normal & Unanswerable 
               & No Para & With Para \\
\midrule
\multicolumn{8}{l}{\textit{Closed-Source Models}} \\
\midrule
GPT4-o                  & 39.63 & 44.49 & 47.74 & 39.71 & 50.13 & 40.04 & 52.29 \\
Gemini-Flash-2.0          & 52.34 & 54.64 & 58.70 & 51.44 & 63.14 & 52.29 & 62.44 \\
Gemini-Flash-1.5        & 43.93 & 49.03 & 51.61 & 47.22 & 40.65 & 44.16 & 57.65 \\
Claude Sonnet 3.5       & 54.63 & 57.42 & 59.30 & 57.63 & 47.98 & 54.33 & 65.29 \\
\midrule
\multicolumn{8}{l}{\textit{Open-Source Models}} \\
\midrule
Qwen-VL2-2B             & 21.20 & 19.41 & 19.93 & 21.02 & 19.24 & 21.16 & 17.59 \\
SmolVLM-2.3B          & 18.88 & 15.15 & 16.36 & 19.99 & 8.49 & 18.30 & 14.65 \\
Phi 3.5-Vision-4B          & 26.15 & 20.96 & 23.12 & 28.72 & 7.66 & 25.12 & 22.19 \\
Qwen-VL2-7B             & 37.18 & 31.61 & 33.43 & 37.13 & 28.99 & 35.30 & 37.47 \\
InternVL2.5-8B            & 36.74 & 35.10 & 32.38 & 31.41 & 53.92 & 35.08 & 39.50 \\
LLama-3.2-Vision-11B  & 23.96 & 26.32 & 31.27 & 29.09 & 9.75 & 25.14 & 27.29 \\

\midrule
\multicolumn{8}{l}{\textit{Chart-Specific Models}} \\
\midrule
ChartGemma              & 7.01 & 4.74 & 8.98 & 8.38 & 0.27 & 6.94 & 6.24 \\
ChartInstruct-LLama2    & 5.97 & 2.84 & 2.48 & 6.03 & 0.0 & 5.64 & 0.0 \\
TinyChart               & 13.75 & 11.20 & 13.69 & 16.28 & 0.27 & 15.25 & 0.38 \\
\bottomrule
\end{tabular}
} %
\end{table}

\subsection{Ablation Studies}
Table \ref{tab:ablation-3-in-1} shows ablation results on\chartqapro{} on three independent dimensions: \emph{(A) Chart Type}, \emph{(B) Answer Type}, and \emph{(C) Paragraph Presence}.
\paragraph{Chart Type:} Closed-source models demonstrate greater robustness to complex visual layouts, such as dashboards and infographics. In contrast, both open-source and chart-specific models exhibit a performance decline on such complex visuals.
\paragraph{Answer Type:} Among closed-source models, GPT-4o and Gemini Flash 2.0 handle unanswerable questions relatively well, while Gemini Flash 1.5 and Claude Sonnet 3.5 show lower robustness. Similarly, open-source models generally perform worse on unanswerable questions. Chart-specific models, however, struggle significantly, with performance near zero, highlighting their limited ability to handle ambiguous or missing information.
\paragraph{Paragraph Presence:} Closed-source models can effectively utilize the additional context. Among open-source models, smaller models struggle with this added context, while larger models are more robust.  Chart-specific models perform poorly with added context, likely due to overfitting, except for ChartGemma \cite{masry2024chartgemma}.

Overall, our analysis shows that while closed-source models generally lack in recognizing data values (visual perception), open-source and chart-specific models struggle with visual complexity, ambiguous information, and added context, highlighting the need for improvements to match closed-source models in chart understanding. We present exemplar details in \ref{app:ablations} and Figure \ref{fig:ablations_error_analysis}.
\section{Conclusion}

We introduced\chartqapro{}, a more diverse and challenging benchmark for chart question answering, designed to push the limits of current vision-language models (VLMs) in real-world chart reasoning. By incorporating 1341 charts from 157 sources and a broad spectrum of question types—including factoid, multiple-choice, fact-checking, conversational, and hypothetical queries—our benchmark reveals significant performance gaps between existing models and human-level understanding. Our extensive evaluation shows that even the strongest closed-source models experience substantial performance drops, underscoring that chart reasoning remains an unsolved challenge. Through detailed error analysis and ablation studies, we identify key areas for improvement, paving the way for future advancements in multimodal reasoning. We hope\chartqapro{} serves as a catalyst for developing more robust and capable models for real-world chart comprehension.

As future work, we plan on expanding the benchmark by introducing dynamic and interactive charts and dashboards, as current benchmarks only use screenshots of the charts -- which often does not happen in real-world scenarios. We also aim to curate a large-scale training dataset in reasoning formats following recent advances in LLM training, hoping to develop significantly more proficient chart understanding and reasoning models.

\section*{Limitations}
While \chartqapro{} is designed to comprehensively evaluate chart understanding, there are a few limitations to consider. First, our benchmark primarily focuses on chart question answering (ChartQA) as the core evaluation task. While this task effectively measures a model’s ability to extract, interpret, and reason over chart data, other chart-related tasks—such as chart-to-summary generation or chart-to-code translation—are also valuable and remain unexplored in this work.

Second, although we carefully tuned prompts to ensure fair and consistent evaluation across all models, performance may vary slightly by applying further prompt engineering techniques. While certain models might benefit from additional prompt engineering, we do not expect such adjustments to lead to substantial improvements or change the overall findings in our study.

Third, the dashboards included in\chartqapro{} are static screenshots rather than interactive elements. In real-world scenarios, most dashboards often allow users to hover, filter, or manipulate data dynamically, which can impact how insights are extracted. Since our benchmark does not incorporate interactivity, models are evaluated solely on the static visual and textual information presented in the images.

Despite these limitations,\chartqapro{} provides a rigorous and diverse benchmark that highlights key challenges in chart reasoning and serves as a valuable resource to advance multimodal research.

\section*{Ethical Considerations}
During the dataset collection process, we carefully considered several ethical aspects to ensure the integrity of our work. All collected images underwent a thorough manual review by the authors to filter out any content that could be considered harmful or offensive. Additionally, our benchmark does not feature any proprietary data, as all charts were sourced from publicly available online platforms. We plan to release the dataset only for research purposes. 

The question-answer (QA) generation process was carried out exclusively by the authors, all of whom are researchers with expertise in chart understanding. While large vision-language models (LVLMs) were used as assistance tools in the QA expansion process, all questions and answers were manually reviewed and refined to ensure accuracy, coherence, and ethical neutrality. No external or paid annotators were involved in this study. Instead, all individuals who contributed to dataset annotation were granted co-authorship to recognize their contributions. All annotators were informed that their annotations would be included in the dataset released for research purposes. Finally, AI writing assistants were used to refine the writing and enhance the paper’s presentation.

\section*{Acknowledgement}
We would like to thank the anonymous reviewers for their helpful feedback. This research was supported by the Natural Sciences Engineering Research Council (NSERC) of Canada and Canada Foundation for Innovation (CFI). Additionally, it received support through a Google Cloud Platform (GCP) credits award from Google's PaliGemma Academic Program. 

\bibliography{chart2text}
\newpage
\appendix
\section{Appendices}
\subsection{Dataset Construction}
\label{app:AMT}
In this section, we outline the sources from which we collected all the chart images.
\paragraph{$\bullet$\, Pew.} The Pew Research Center \cite{pewresearch} publishes data reports on social issues, public opinion, and demographic trends, often using charts and text to tell a clear data story. For our dataset, we collected a subset of images from a larger corpus compiled by \cite{islam-etal-2024-datanarrative}.  This corpus, which includes 22,760 figures (charts and other images) scraped from the Pew Research website up to March 14, 2024, provided our initial pool of images.  From this pool, we selected a subset and then further filtered it.  We excluded simple statistical charts and basic visualizations like single bar or line charts, focusing instead on visually diverse charts covering a range of topics. We further collected the paragraphs associated with these chart images. The associated paragraphs not only describe the visualized data but also offer additional context not explicitly mentioned in the charts, enhancing their interpretive value.

\paragraph{$\bullet$\, Tableau.} We used Tableau Public \cite{tableaupublic} as a source for our dataset. Tableau Public allows users to create and share interactive dashboards made up of data visualizations on a variety of topics. We sourced the chart images for our dataset from a larger corpus collected and curated by \cite{islam-etal-2024-datanarrative}. Due to the complex nature of the dashboard representation, they manually curated the data, focusing on dashboards with stories presented in a paginated format, where each page included both text and a corresponding chart. The final Tableau corpus from \cite{islam-etal-2024-datanarrative} consists of 100 dashboards covering a diverse range of topics and chart images. From this pool, we manually selected our own Tableau corpus based on specific criteria. We ensured that the selected dashboards included a variety of chart images, accompanying paragraphs of reasonable length, and a broad representation of topics.

\paragraph{$\bullet$\, OWID.} Our World in Data (OWID) \cite{owid} is a non-profit online platform that provides research and data on a wide range of global issues, including poverty, disease, hunger, climate change, and inequality. We sourced chart images from OWID focusing on including a diverse range of complex charts, i.e., multi-series line charts and multi-column bar charts to enhance the dataset.

\paragraph{$\bullet$\, PPIC.} The Public Policy Institute of California (PPIC) \cite{ppic} is an independent research institute dedicated to informing public policy in California. Through data-driven research and analysis, PPIC examines a wide range of policy areas, including the economy, education, environment, and governance. Similar to OWID corpus we sourced chart images that excluded simple statistical charts and basic visualizations like single bar or line charts, focusing instead on visually diverse charts covering a range of topics to enhance the dataset.

\paragraph{$\bullet$\, WebCharts.} 
We built WebCharts corpus by leveraging prior work from efforts from ChartGemma \cite{masry2024chartgemma} and ChartInstruct \cite{masry2024chartinstruct}. Their chart image collection process began with a seed list of 157 websites known to host charts (originally compiled by \citet{hoque-vis-search-2019}), then querying Google Images using terms like ``chart images'', ``graphs'', and ``visual data.'' This initial search yielded a large number of images, which we then filtered using a binary Vision Transformer (ViT) \cite{vit} classifier to identify and isolate chart images.  Any remaining non-chart images were manually removed to ensure accuracy.  This process, starting with the seed list and refined through image search and classification, ultimately gave us a pool of 41,000 chart images.  From this larger set, we carefully selected 800 charts, prioritizing visual and topical diversity.  Our final selection emphasizes high visual quality and representation across a range of chart styles, formats, and subject matter. In addition, we manually curated 200 infographic charts, which serve to highlight data visualization trends aimed at storytelling and public engagement.

\begin{figure*}[t]
    \includegraphics[width=\textwidth]{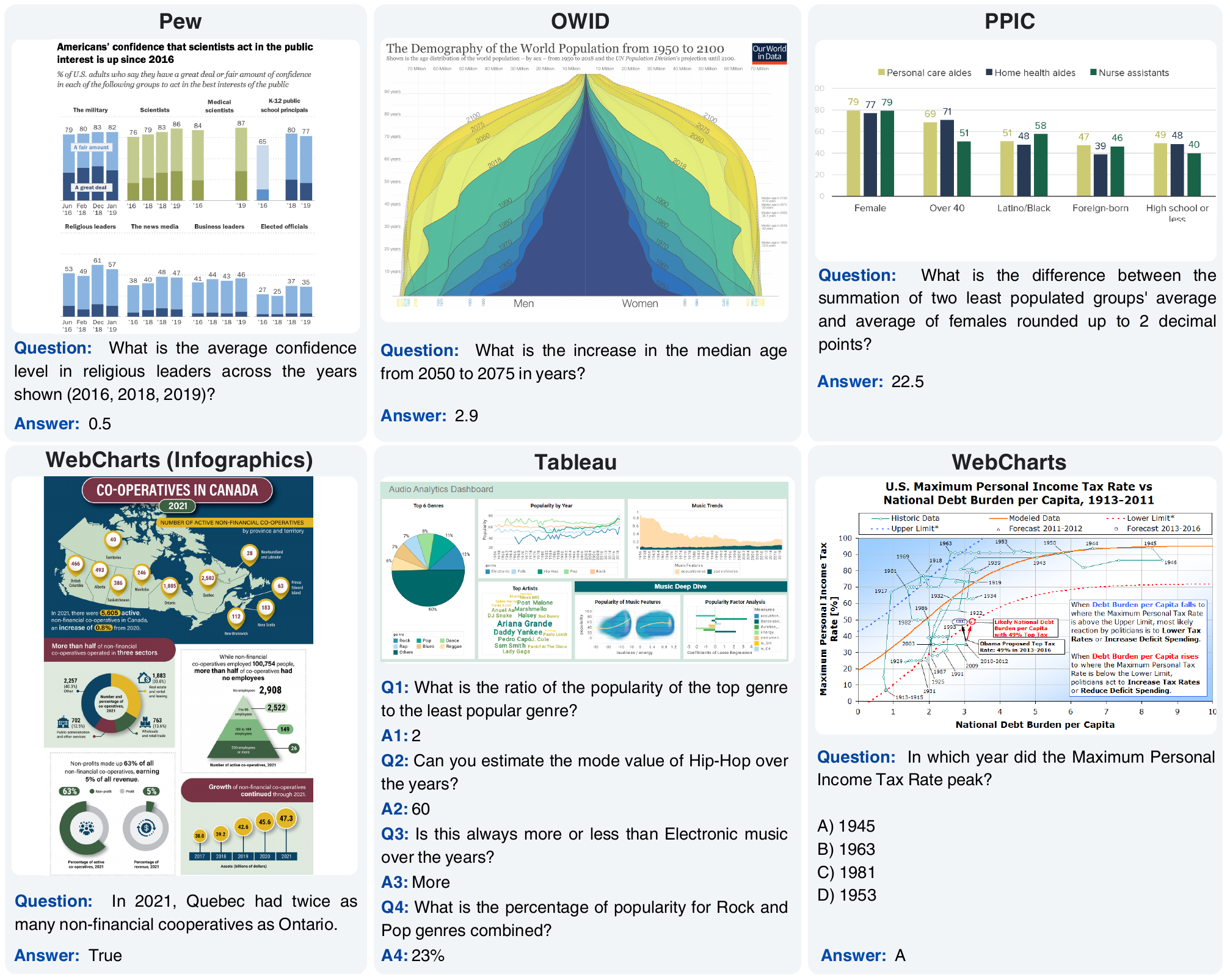}
    \caption{Example of chart images collected from different sources and their corresponding QA pairs in \chartqapro{}.}
    \label{fig:src_charts}
\end{figure*}

\begin{figure*}[t]
    \includegraphics[width=\textwidth]{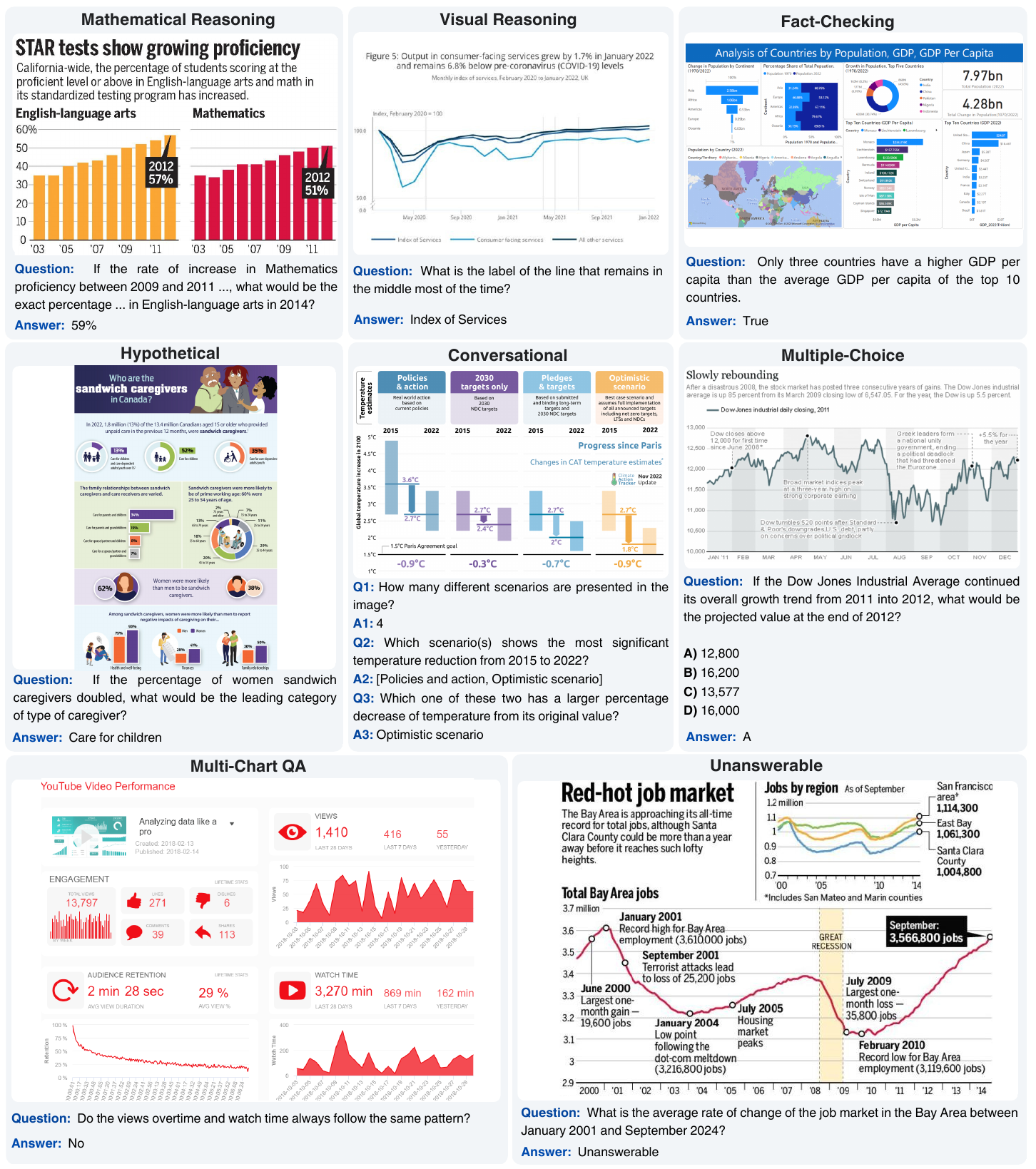}
    \caption{More examples of different question types in \chartqapro{}.}
    \label{fig:moreques_type}
\end{figure*}

The extensive coverage of our dataset stands in contrast to prior datasets, which often relied on a limited number of sources, such as Statista \cite{statista} or Pew \cite{pewresearch}, and exhibited restricted stylistic variation. By incorporating a significantly larger pool of sources, our dataset ensures broader domain coverage and richer stylistic representation, addressing critical limitations in existing chart corpora. In addition to collecting the chart images, we also gathered metadata associated with them, including the URL, alt text, and other relevant details. Finally, the careful curation process resulted in a diverse collection of 1341 chart images spanning various types and styles. We provide samples from each source in Figure \ref{fig:src_charts} and our different questions categories in Figure \ref{fig:moreques_type}.

\subsection{Complex Visualizations}
\label{app:comlpex_vis}

Multi-chart images, infographics, and dashboards all vital data visualizations that serve different purposes. Multi-chart images combine multiple charts in a single visual often for comparison or to present different aspects of a dataset. Infographics integrate text, images, and charts to explain concepts or tell a story, focusing on clarity and engagement rather than detailed data analysis. Dashboards organize charts, tables, and key metrics in a structured layout, providing an overview of important data for quick interpretation and decision-making. Table \ref{tab:multi_chart_examples} presents examples of each type for reference.

\definecolor{lightblue}{RGB}{230, 245, 255}
\definecolor{lightgreen}{RGB}{230, 255, 230}
\definecolor{lightyellow}{RGB}{255, 250, 220}

\begin{table*}[h]
    \centering
    \renewcommand{\arraystretch}{1.5} 
    \setlength{\tabcolsep}{12pt} 
    \captionsetup{font=small, labelfont=bf}
    \arrayrulecolor{black} 
    
    \begin{tabular}{>{\columncolor{lightblue}}c >{\columncolor{lightgreen}}c}
        \toprule
        \textbf{Multi-Chart Image} & \textbf{Infographic} \\
        \midrule
        \cellcolor{lightblue} \includegraphics[width=0.42\textwidth]{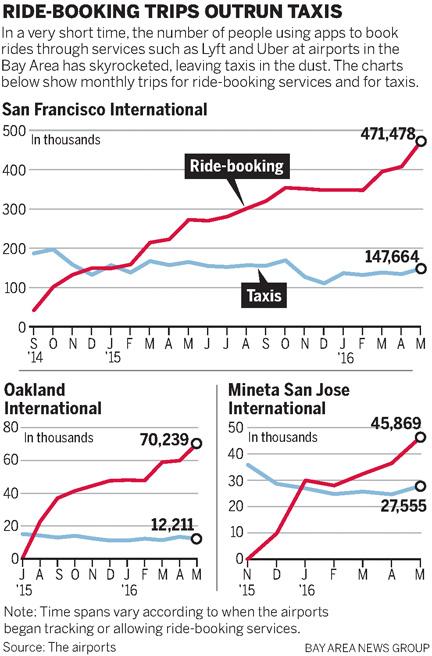} &
        \cellcolor{lightgreen} \includegraphics[width=0.42\textwidth]{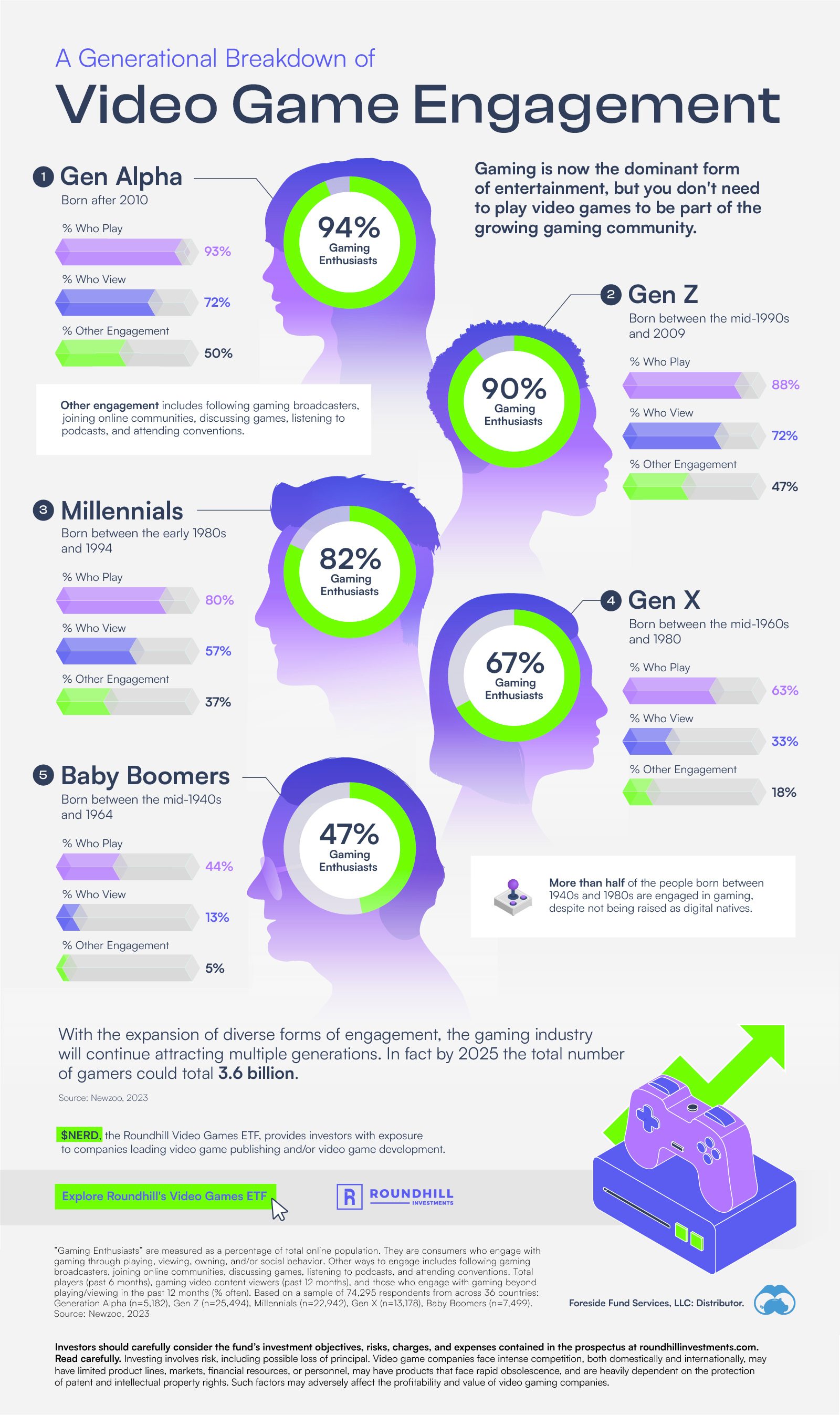} \\
        \midrule
        \cellcolor{lightblue} Combines multiple charts to compare data &
        \cellcolor{lightgreen} Integrates text and visuals to tell a story \\
        \bottomrule
    \end{tabular}

    \vspace{1em} 

    \begin{tabular}{>{\columncolor{lightyellow}}c}
        \toprule
        \textbf{Dashboard} \\
        \midrule
        \cellcolor{lightyellow} \includegraphics[width=0.85\textwidth]{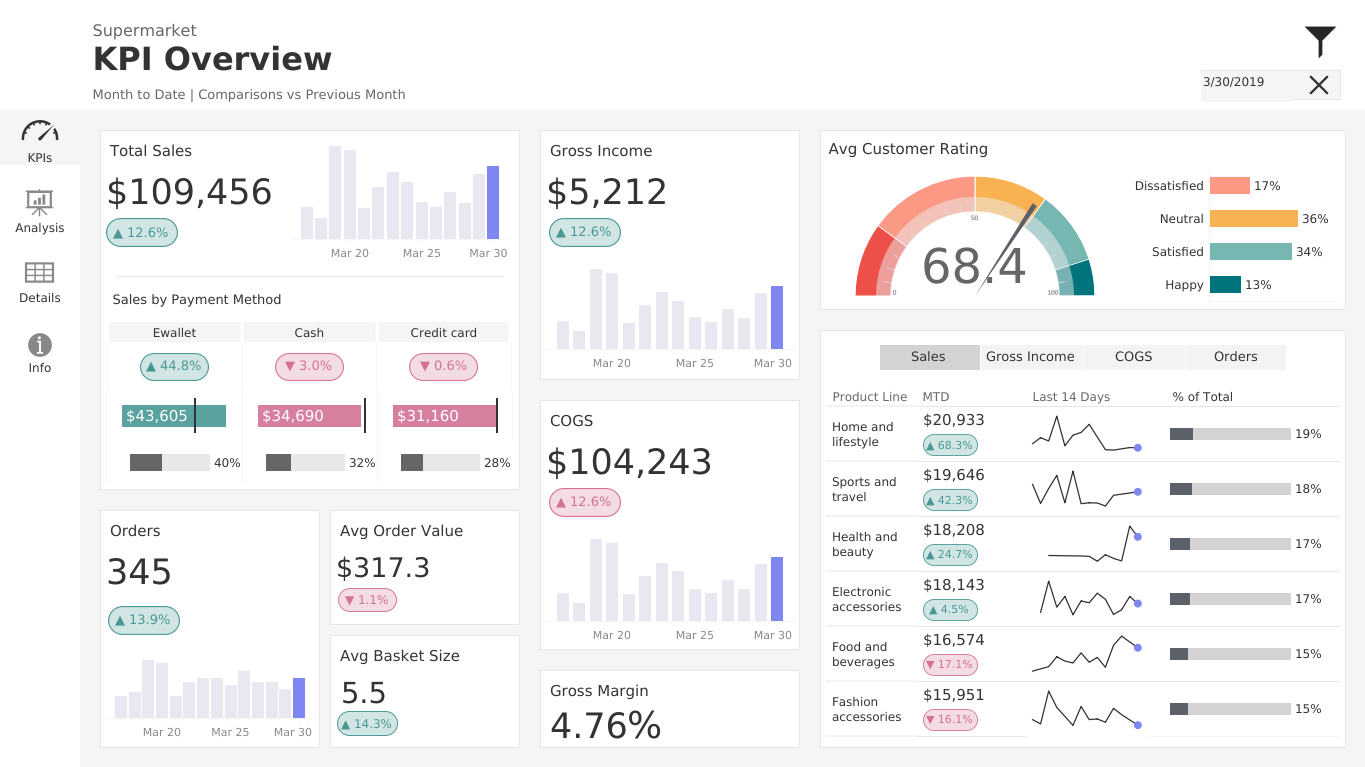} \\
        \midrule
        \cellcolor{lightyellow} Displays key metrics for quick interpretation \\
        \bottomrule
    \end{tabular}

    \caption{Examples of Multi-Chart Images, Infographics, and Dashboards, with distinct background colors for clarity.}
    \label{tab:multi_chart_examples}
\end{table*}

\subsection{Dataset Analysis}
\label{app:data_analysis}

\subsubsection{Visual Diversity}
\label{app:visual_diversity}
\begin{figure*}[t]
    \includegraphics[width=\textwidth]{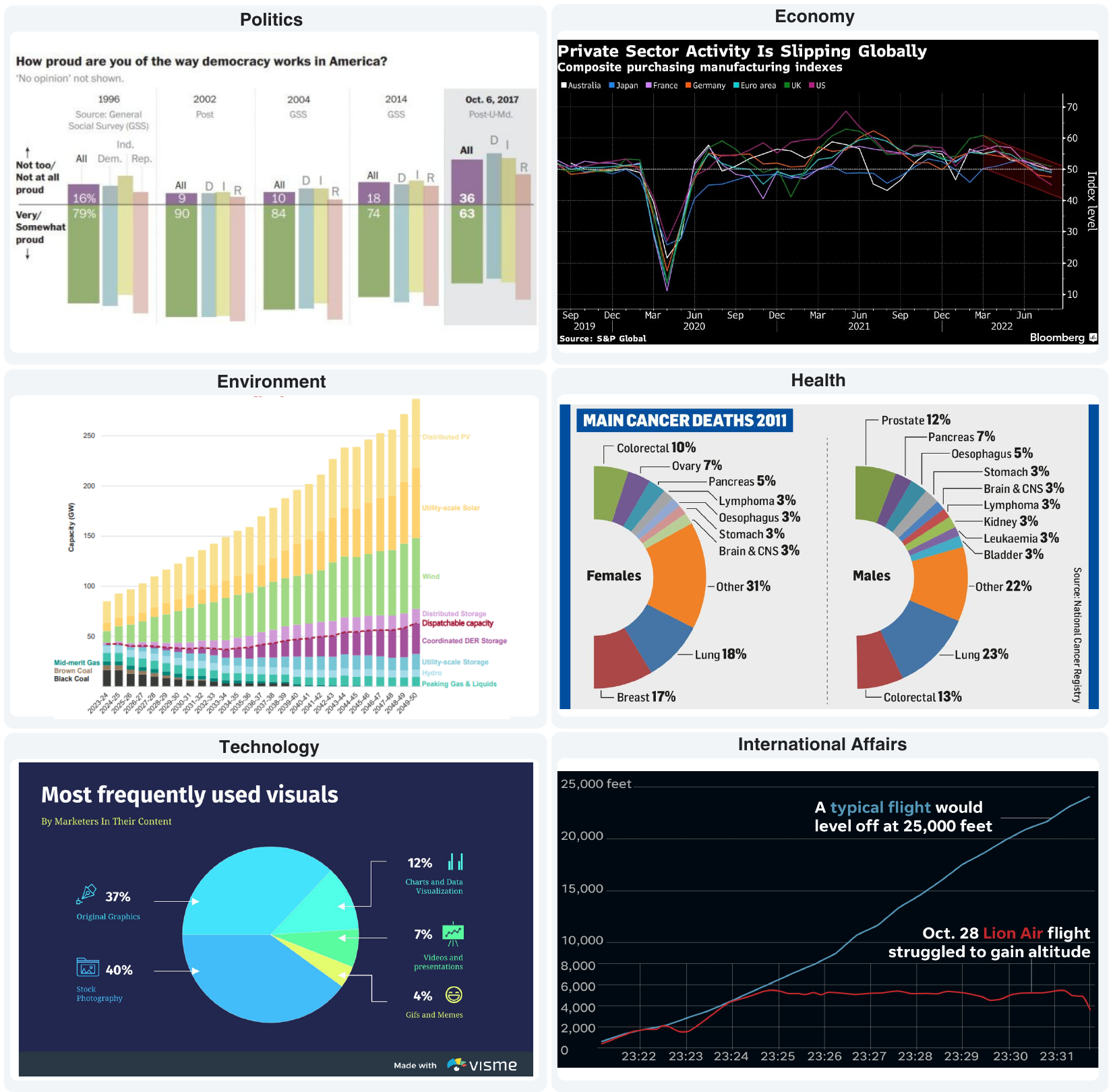}
    \caption{Examples of different charts related to major topics, i.e., `Politics', `Environment', `Economy', `Health', `Technology', `International Affairs' etc. in \chartqapro{}.}
    \label{fig:topic_charts}
\end{figure*}

Figure \ref{fig:topic_charts} shows example charts from diverse topics in our \chartqapro{} benchmark.

\begin{figure*}[t]
    \includegraphics[width=\textwidth]{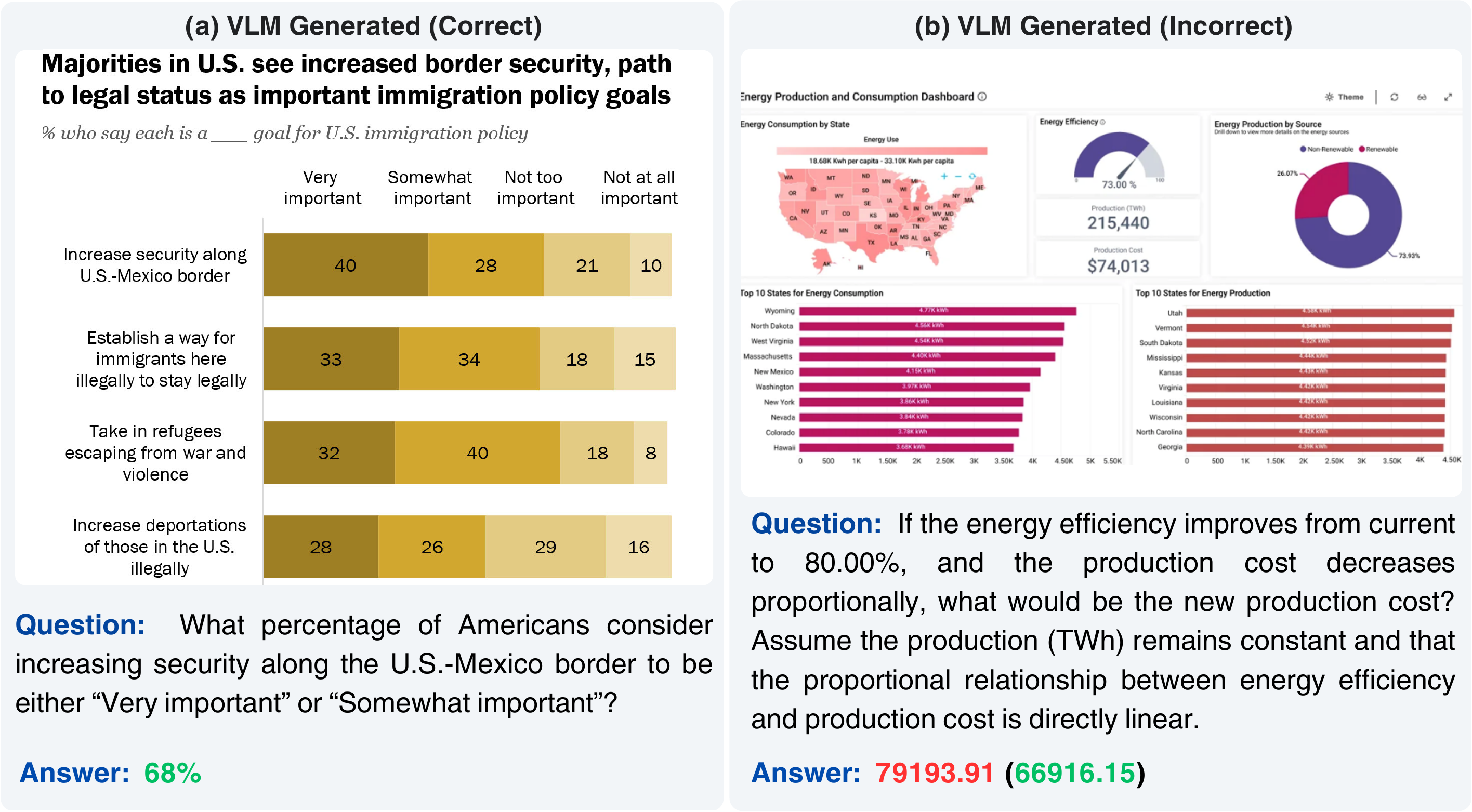}
    \caption{Examples of VLM-assisted question-and-answer pairs, where: (a) the VLM generates a question along with a correct answer, marked in \textcolor{green}{Green} text, (b) the VLM generates a question, but the answer is incorrect, marked in \textcolor{red}{Red} text.}
    \label{fig:llm_gen}
\end{figure*}

 \begin{figure}[t]
    \includegraphics[width=\columnwidth]{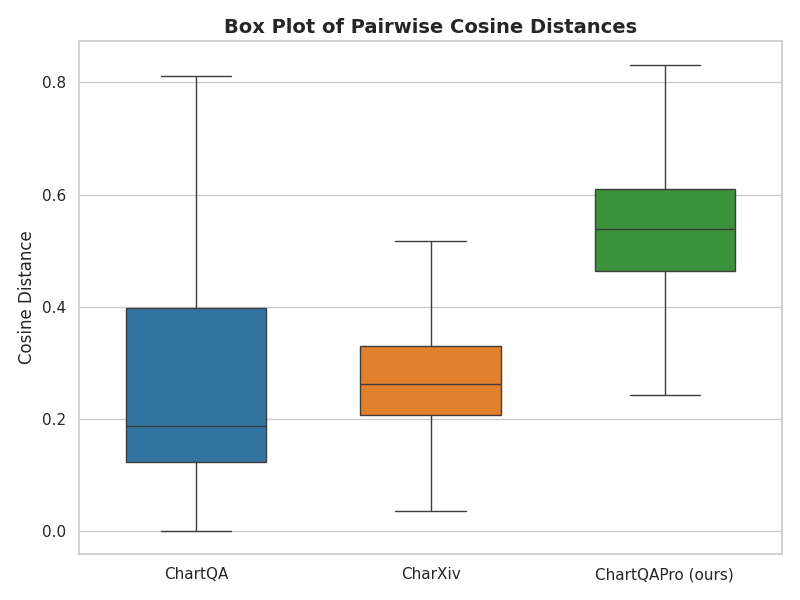}
    \caption{Box plot of pairwise cosine distances among chart images. \textsc{ChartQAPro} exhibits a higher median and consistently larger distances, indicating significantly greater visual diversity.}
    \label{fig:visual_diversity}
\end{figure}

\subsubsection{Linguistic Diversity}
\label{app:linguistic_diversity}
In our analysis, we first quantified the lexical diversity of each dataset by computing the Type-Token Ratio (TTR). Let \(T\) denote the total number of tokens (i.e., words) extracted from a dataset and \(U\) the number of unique tokens. The TTR is given by 
\[
\text{TTR} = \frac{U}{T}.
\]
Higher TTR values indicate a richer vocabulary and, consequently, greater lexical diversity. Our experiments revealed that the ChartQAPro dataset achieved a TTR of 0.1516, compared to 0.1377 for ChartQA and 0.1189 for Chartxiv.

To assess semantic diversity, we computed the average pairwise cosine distance between text embeddings. We obtained vector representations for each text using the Sentence-BERT model \texttt{all-MiniLM-L6-v2}. For a given text sample \(i\), let \(\mathbf{v}_i\) denote its embedding. The cosine distance between two embeddings \(\mathbf{v}_i\) and \(\mathbf{v}_j\) is calculated as 
\[
d(\mathbf{v}_i, \mathbf{v}_j) = 1 - \frac{\mathbf{v}_i \cdot \mathbf{v}_j}{\|\mathbf{v}_i\|\,\|\mathbf{v}_j\|}.
\]
We then computed the overall semantic diversity as the average of these distances over all unique pairs,
\[
D_{\text{avg}} = \frac{2}{N(N-1)} \sum_{i<j} d(\mathbf{v}_i, \mathbf{v}_j),
\]
where \(N\) is the total number of text samples. A higher value of \(D_{\text{avg}}\) indicates that the texts are more semantically dispersed. ChartQAPro showed an average cosine distance of 0.8439, compared to 0.7558 for ChartQA and 0.7831 for Chartxiv.

\begin{figure*}[t]
    \includegraphics[width=\textwidth]{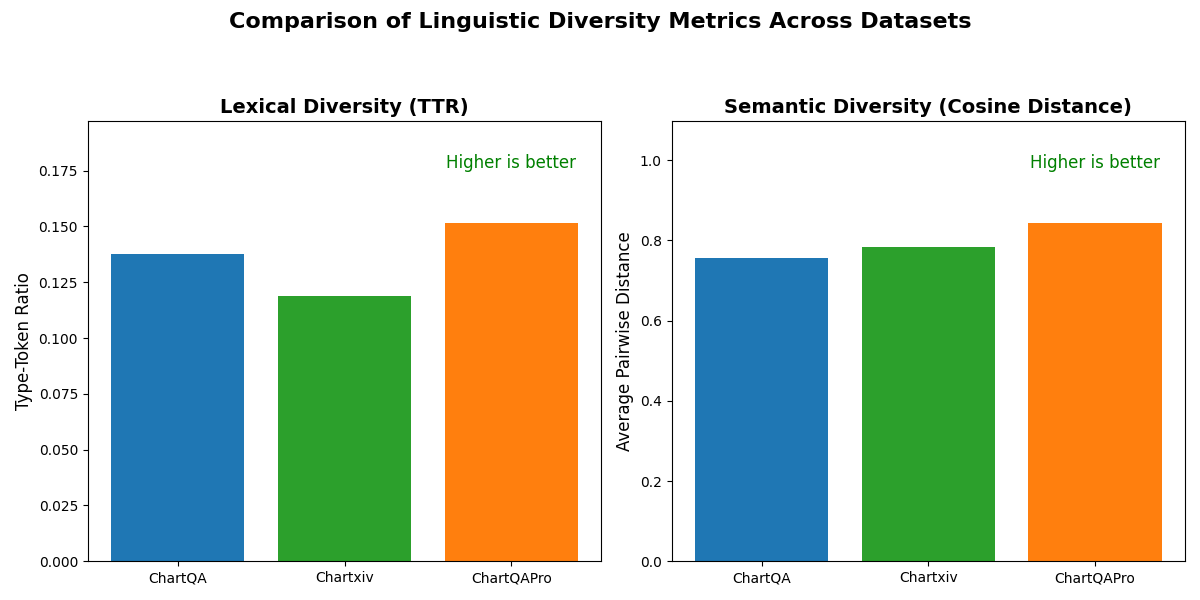}
    \caption{Linguistic Diversity Comparison Across Datasets. The figure shows lexical diversity (TTR) and semantic diversity (cosine distance) for ChartQA, Chartxiv, and ChartQAPro. Higher TTR and semantic diversity indicate richer vocabulary and broader semantic coverage. ChartQAPro exhibits the highest diversity.}
    \label{fig:linguistic_diversity}
\end{figure*}

Overall, these metrics—lexical diversity (TTR) and semantic diversity (average pairwise cosine distance computed using Sentence-BERT \texttt{all-MiniLM-L6-v2})—demonstrate that the ChartQAPro dataset is linguistically more diverse than the previous benchmarks. Figure~\ref{fig:linguistic_diversity} illustrates these findings, showing that ChartQAPro outperforms ChartQA and Chartxiv with higher TTR and semantic diversity.

\subsection{Prompts for Models Evaluation}
\label{app:prompts-evaluation}

To promote transparency and reproducibility, we provide the exact prompts used to evaluate our models. Table \ref{tab:prompt_templates_direct} presents the prompts for the Direct Question Answering setup, Table \ref{tab:prompt_templates_cot} details those for the Chain-of-Thought setup, and Table \ref{tab:prompt_templates_pot} outlines the prompts for the Program-of-Thought setup.

\begin{table*}[htbp]
  \centering
  \caption{Prompt Templates for Each Question Category in the Direct setup.}
  \label{tab:prompt_templates_direct}
  \rowcolors{2}{gray!10}{white}
  \begin{tabularx}{\textwidth}{>{\bfseries}l X}
    \toprule
    Category & Prompt Template \\ 
    \midrule
    Factoid & 
    \begin{minipage}[t]{\linewidth}
      \small
      You are given a factoid question that you need to answer based on the provided image.\\[1ex]
      Your answer should be a single word, number, or phrase. If the question is unanswerable based on the information in the provided image, your answer should be unanswerable. Do not generate units. But if numerical units such as \texttt{million}, \texttt{m}, \texttt{billion}, \texttt{B}, or \texttt{K} are required, use the exact notation shown in the chart.\\[1ex]
      If there are multiple answers, put them in brackets using this format \texttt{['Answer1', 'Answer2']}.\\[1ex]
      Remember to generate the final answer only without any additional text!\\[1ex]
      Question: \textless question\textgreater
    \end{minipage} \\
    \midrule
    Multi Choice & 
    \begin{minipage}[t]{\linewidth}
      \small
      You are given a question along with different possible answers. You need to select the correct answer from them based on the provided image.\\[1ex]
      Your answer should be one of the options letters only: \texttt{a}, \texttt{b}, \texttt{c} or \texttt{d} (just the letter itself without any additional text). If the question is unanswerable based on the information in the provided image, your answer should be unanswerable.\\[1ex]
      If there are multiple answers, put them in brackets using this format \texttt{['Answer1', 'Answer2']}.\\[1ex]
      Remember to generate the final answer only without any additional text!\\[1ex]
      Question: \textless question\textgreater
    \end{minipage} \\
    \midrule
    Hypothetical & 
    \begin{minipage}[t]{\linewidth}
      \small
      You are given a hypothetical question that you need to answer based on the provided image.\\[1ex]
      Your answer should be a single word, number, or phrase. If the question is unanswerable based on the information in the provided image, your answer should be unanswerable. Do not generate units. But if numerical units such as \texttt{million}, \texttt{m}, \texttt{billion}, \texttt{B}, or \texttt{K} are required, use the exact notation shown in the chart.\\[1ex]
      If there are multiple answers, put them in brackets using this format \texttt{['Answer1', 'Answer2']}.\\[1ex]
      Remember to generate the final answer only without any additional text!\\[1ex]
      Question: \textless question\textgreater
    \end{minipage} \\
    \midrule
    Fact Checking & 
    \begin{minipage}[t]{\linewidth}
      \small
      You are given a fact statement that you need to assess based on the provided image.\\[1ex]
      Your answer should be either \texttt{true} or \texttt{false} (without any additional text). If the question is unanswerable based on the information in the provided image, your answer should be unanswerable.\\[1ex]
      If there are multiple answers, put them in brackets using this format \texttt{['Answer1', 'Answer2']}.\\[1ex]
      Remember to generate the final answer only without any additional text!\\[1ex]
      Question: \textless question\textgreater
    \end{minipage} \\
    \midrule
    Conversational & 
    \begin{minipage}[t]{\linewidth}
      \small
      You are given a multi-turn conversation, and your job is to answer the final question based on the conversation history and the information in the provided image.\\[1ex]
      Your answer should be a single word, number, or phrase. If the question is unanswerable based on the information in the provided image, your answer should be unanswerable. Do not generate units. But if numerical units such as \texttt{million}, \texttt{m}, \texttt{billion}, \texttt{B}, or \texttt{K} are required, use the exact notation shown in the chart.\\[1ex]
      If there are multiple answers, put them in brackets using this format \texttt{['Answer1', 'Answer2']}.\\[1ex]
      Remember to generate the final answer only without any additional text!\\[1ex]
      Conversation: \textless conversation\textgreater \quad
      Question: \textless question\textgreater
    \end{minipage} \\
    \bottomrule
  \end{tabularx}
\end{table*}

\begin{table*}[htbp]
  \centering
  \caption{Prompt Templates for Each Question Category under the Chain of Thought Setup}
  \label{tab:prompt_templates_cot}
  \rowcolors{2}{gray!10}{white}
  \begin{tabularx}{\textwidth}{>{\bfseries}l X}
    \toprule
    Category & Prompt Template \\ 
    \midrule
    Factoid &
    \begin{minipage}[t]{\linewidth}
      \small
      You are given a factoid question that you need to answer based on the provided image.\\[1ex]
      You need to think step-by-step, but your final answer should be a single word, number, or phrase. If the question is unanswerable based on the information in the provided image, your answer should be unanswerable. Do not generate units. But if numerical units such as million, m, billion, B, or K are required, use the exact notation shown in the chart.\\[1ex]
      If there are multiple final answers, put them in brackets using this format ['Answer1', 'Answer2']. . Remember to think step-by-step and format the final answer in a separate sentence like "The answer is X"\\[1ex]
      Question: \textless question\textgreater
    \end{minipage} \\
    \midrule
    Multi Choice &
    \begin{minipage}[t]{\linewidth}
      \small
      You are given a question along with different possible answers. You need to select the correct answer from them based on the provided image.\\[1ex]
      You need to think step-by-step, but your final answer should be one of the options letters only: a, b, c or d (just the letter itself without any additional text). If the question is unanswerable based on the information in the provided image, your answer should be unanswerable.\\[1ex]
      If there are multiple final answers, put them in brackets using this format ['Answer1', 'Answer2']. . Remember to think step-by-step and format the final answer in a separate sentence like "The answer is X"\\[1ex]
      Question: \textless question\textgreater
    \end{minipage} \\
    \midrule
    Hypothetical &
    \begin{minipage}[t]{\linewidth}
      \small
      You are given a hypothetical question that you need to answer based on the provided image.\\[1ex]
      You need to think step-by-step, but your final answer should be a single word, number, or phrase. If the question is unanswerable based on the information in the provided image, your answer should be unanswerable. Do not generate units. But if numerical units such as million, m, billion, B, or K are required, use the exact notation shown in the chart.\\[1ex]
      If there are multiple final answers, put them in brackets using this format ['Answer1', 'Answer2']. Remember to think step-by-step and format the final answer in a separate sentence like "The answer is X"\\[1ex]
      Question: \textless question\textgreater
    \end{minipage} \\
    \midrule
    Fact Checking &
    \begin{minipage}[t]{\linewidth}
      \small
      You are given a fact statement that you need to assess based on the information in the provided image.\\[1ex]
      You need to think step-by-step, but your final answer should be either true or false (without any additional text). If the question is unanswerable based on the information in the provided image, your answer should be unanswerable.\\[1ex]
      If there are multiple final answers, put them in brackets using this format ['Answer1', 'Answer2']. . Remember to think step-by-step and format the final answer in a separate sentence like "The answer is X"\\[1ex]
      Question: \textless question\textgreater
    \end{minipage} \\
    \midrule
    Conversational &
    \begin{minipage}[t]{\linewidth}
      \small
      You are given a multi-turn conversation, and your job is to answer the final question based on the conversation history and the information in the provided image.\\[1ex]
      You need to think step-by-step, but your final answer should be a single word, number, or phrase. If the question is unanswerable based on the information in the provided image, your answer should be unanswerable. Do not generate units. But if numerical units such as million, m, billion, B, or K are required, use the exact notation shown in the chart.\\[1ex]
      If there are multiple final answers, put them in brackets using this format ['Answer1', 'Answer2']. . Remember to think step-by-step and format the final answer in a separate sentence like "The answer is X"\\[1ex]
      Conversation: \textless conversation\textgreater \quad
      Question: \textless question\textgreater
    \end{minipage} \\
    \bottomrule
  \end{tabularx}
\end{table*}

\begin{table*}[htbp]
  \centering
  \caption{Prompt Templates for Each Question Category in the Program-of-Thought setup.}
  \label{tab:prompt_templates_pot}
  \rowcolors{2}{gray!10}{white}
  \begin{tabularx}{\textwidth}{>{\bfseries}l X}
    \toprule
    Category & Prompt Template \\
    \midrule
    Factoid &
    \begin{minipage}[t]{\linewidth}
      \small
      You are given a factoid question that you need to answer based on the provided image.\\[1ex]
      You need to write an executable python code that calculates and prints the final answer, but your final answer should be a single word, number, or phrase. If the question is unanswerable based on the information in the provided image, your answer should be unanswerable. Do not generate units. But if numerical units such as million, m, billion, B, or K are required, use the exact notation shown in the chart.\\[1ex]
      If there are multiple final answers, put them in brackets using this format ['Answer1', 'Answer2'].\\[1ex]
      Remember to return a python code only without any additional text.\\[1ex]
      Question: \textless question\textgreater
    \end{minipage} \\
    \midrule
    Multi Choice &
    \begin{minipage}[t]{\linewidth}
      \small
      You are given a question along with different possible answers. You need to select the correct answer from them based on the provided image.\\[1ex]
      You need to write an executable python code that calculates and prints the final answer, but your final answer should be one of the options letters only: a, b, c or d (just the letter itself without any additional text). If the question is unanswerable based on the information in the provided image, your answer should be unanswerable.\\[1ex]
      If there are multiple final answers, put them in brackets using this format ['Answer1', 'Answer2'].\\[1ex]
      Remember to return a python code only without any additional text.\\[1ex]
      Question: \textless question\textgreater
    \end{minipage} \\
    \midrule
    Hypothetical &
    \begin{minipage}[t]{\linewidth}
      \small
      You are given a hypothetical question that you need to answer based on the provided image.\\[1ex]
      You need to write an executable python code that calculates and prints the final answer, but your final answer should be a single word, number, or phrase. If the question is unanswerable based on the information in the provided image, your answer should be unanswerable. Do not generate units. But if numerical units such as million, m, billion, B, or K are required, use the exact notation shown in the chart.\\[1ex]
      If there are multiple final answers, put them in brackets using this format ['Answer1', 'Answer2'].\\[1ex]
      Remember to return a python code only without any additional text.\\[1ex]
      Question: \textless question\textgreater
    \end{minipage} \\
    \midrule
    Fact Checking &
    \begin{minipage}[t]{\linewidth}
      \small
      You are given a fact statement that you need to assess based on the information in the provided image.\\[1ex]
      You need to write an executable python code that calculates and prints the final answer, but your final answer should be either true or false (without any additional text). If the question is unanswerable based on the information in the provided image, your answer should be unanswerable.\\[1ex]
      If there are multiple final answers, put them in brackets using this format ['Answer1', 'Answer2'].\\[1ex]
      Remember to return a python code only without any additional text.\\[1ex]
      Question: \textless question\textgreater
    \end{minipage} \\
    \midrule
    Conversational &
    \begin{minipage}[t]{\linewidth}
      \small
      You are given a multi-turn conversation, and your job is to answer the final question based on the conversation history and the information in the provided image.\\[1ex]
      You need to write an executable python code that calculates and prints the final answer, but your final answer should be a single word, number, or phrase. If the question is unanswerable based on the information in the provided image, your answer should be unanswerable. Do not generate units. But if numerical units such as million, m, billion, B, or K are required, use the exact notation shown in the chart.\\[1ex]
      If there are multiple final answers, put them in brackets using this format ['Answer1', 'Answer2'].\\[1ex]
      Remember to return a python code only without any additional text.\\[1ex]
      Conversation: \textless conversation\textgreater \quad
      Question: \textless question\textgreater
    \end{minipage} \\
    \bottomrule
  \end{tabularx}
\end{table*}

\begin{table*}[htbp]
  \centering
  \caption{Prompt Templates for generating questions using VLMs.}
  \label{tab:prompt_templates_ques_gen}
  \rowcolors{2}{gray!10}{white}
  \begin{tabularx}{\textwidth}{>{\bfseries}l X}
    \toprule
    Category & Prompt Template \\ 
    \midrule
    Reasoning & 
    \begin{minipage}[t]{\linewidth}
      \small
        Generate some of the most difficult Factoid Questions alongside the Corresponding Answers for the given image.\\[1ex]
    The questions could be related to numerical or visual reasoning. And the Answers could be a number, text label, or a common phrase (Yes, No).\\[1ex]
    You should respond in an Array of JSON objects format with the following keys: (i) Question, and (ii) Answer.
    \end{minipage} \\
    \midrule
    Multiple-Choice & 
    \begin{minipage}[t]{\linewidth}
      \small
      I will upload some charts, graphs, infographics or other data visualizations. Generate five multiple-choice questions. \\[1ex]
      Each question should contain four options and one correct answer. \\[1ex]
      Questions should require some complex calculations such as trend analysis, anomaly detection, extrapolation, or time series analysis. \\[1ex]
      For the correct answer, show your calculations as well.
    \end{minipage} \\
    \midrule
    Hypothetical & 
    \begin{minipage}[t]{\linewidth}
      \small
        You are an AI that generates concise and specific hypothetical questions based on chart images. Your task is to analyze the chart and generate a short, data-driven hypothetical question that explores future trends, impacts, or extrapolations based on the data. \\[1ex]
        Avoid adding unnecessary explanations or context like `Based on the chart data\dots' or `A meaningful hypothetical question could be\dots'. \\[1ex]
        Keep the question focused and directly related to the chart. The question should make an assumption about future trends, impacts, or extrapolations based on the data.
    \end{minipage} \\
    \midrule
    Fact-Checking & 
    \begin{minipage}[t]{\linewidth}
      \small
        \textbf{\#\#\# Task Description:}\\[1ex]
        Given a chart image in the input, your task is the following: \\[1ex]
        1. Analyze the given chart image and generate `3' to `5' pairs of claims and verdicts about its data. Half of the claims should be supported by the chart’s data, while the other half are refuted.\\[1ex]
        2. Avoid using terms like `rows', `columns', or `elements' from the data table; refer to `chart' or `chart image' instead. If the claim is supported, the verdict should be `True'. If the claim is refuted, the verdict should be `False', followed by a brief explanation.\\[1ex]
        3. The claims should cover comparisons of values or trends, basic statistical values (maximum, minimum, mean, median, mode) without using exact numbers from the chart.\\[1ex]
        4. Ensure a diverse range of claims addressing various visual aspects of the chart, resulting in 3-5 turns of claims and verdicts.\\[1ex]
        5. Generate the claims in between `\textless claim \textgreater' tags, and the verdicts/answers in between `\textless answer \textgreater' tags, without any additional explanation.
    \end{minipage} \\
    \midrule
    Conversational & 
    \begin{minipage}[t]{\linewidth}
      \small
        Show me conversational question answering for analyzing the \textless chart type \textgreater. Make sure this looks like a proper conversation that makes references to previous questions/answers. \\[1ex]
        Make sure all the questions are such that the answer is concise and all questions require arithmetic and logical reasoning. \\[1ex]
        Please make sure to ask mathematical and visual reasoning questions that require multiple complex operations (e..g, `sum', `min', `max', `diff', `ratio', \dots etc).
    \end{minipage} \\
    \bottomrule
  \end{tabularx}
\end{table*}

\subsection{Evaluation Metric}
\label{app:eval-metric}

We evaluate ChartQA model predictions using a \emph{relaxed correctness} metric that handles numeric, textual, and list-based responses through three cases:

\begin{enumerate}

  \item \textbf{MCQ \& Fact Checking Answers:} We use exact match to evaluate these two types of questions.
  
  \item \textbf{Numeric Answers:} For numeric answers (excluding years), a small relative error is allowed. Let $t$ and $p$ denote the target and predicted numbers, respectively. The relative error is defined as
  \[
    E = \frac{|p - t|}{|t|}.
  \]
  The prediction is deemed correct if
  \[
    E \leq \epsilon, \quad \text{with } \epsilon = 0.05.
  \]
  
  \item \textbf{Year Answers:} For answers representing years, an exact match is required to prevent false positives (e.g., 2009 and 2010 would otherwise yield an error rate below 0.05).
  
  \item \textbf{Textual Answers:} For non-numeric textual answers, we use the Average Normalized Levenshtein Similarity (ANLS) metric \cite{biten2019scenetextvisualquestion} rather than strict matching.
\end{enumerate}

A single target--prediction pair is evaluated by the function $C(t,p)$:
\begin{equation}
\resizebox{0.95\linewidth}{!}{$
  C(t, p) =
  \begin{cases}
    ExactM(p, t), & \text{if } \text{\small question is MCQ or Fact Checking}, \\[1ex]
    ExactM(p, t), & \text{if } t \text{ and } p \text{ are years}, \\[1ex]
    1, & \text{if } t \text{ and } p \text{ are numeric and } \dfrac{|p-t|}{|t|} \le 0.05, \\[1ex]
    0, & \text{if } t \text{ and } p \text{ are numeric and } \dfrac{|p-t|}{|t|} > 0.05, \\[1ex]
    \mathrm{ANLS}(p, t), & \text{otherwise.}
  \end{cases}
$}
\end{equation}

\paragraph{List-based Answers:} For responses provided as lists (encoded as strings), we first parse the lists and then compute the score for each corresponding target--prediction pair. Let
\[
  T = [t_1, t_2, \ldots, t_N] \quad \text{and} \quad P = [p_1, p_2, \ldots, p_N].
\]
The overall score for the list is
\begin{equation}
  C_{\text{list}}(T,P) = \frac{1}{N}\sum_{i=1}^{N} C(t_i, p_i).
\end{equation}

\paragraph{Overall Evaluation:} The final accuracy is computed by averaging the scores over all $M$ examples:
\[
  \text{Accuracy} = \frac{1}{M} \sum_{j=1}^{M} C_j.
\]

This metric tolerates minor numeric errors, enforces exact matching for years to avoid misleading correctness from near-miss values, and uses the ANLS score \cite{biten2019scenetextvisualquestion} to assign partial credit for nearly correct textual answers (e.g., ``Female'' vs. ``Females''). We will open-source the evaluation metric code to ensure reproducibility and facilitate further research.

\subsection{Human Baseline Setup}
\label{app:human_baseline}
To approximate an upper bound on model performance, we conducted a human baseline experiment. An expert in‐house graduate student answered 50 randomly sampled questions from each category (Factoid, Conversational, etc.) using the exact same prompts provided to the models to ensures consistency and fairness. The resulting accuracies are reported in Table \ref{tab:prompt-question-accuracy} under the Direct prompting setup, as Chain‐of‐Thought and Program‐of‐Thought formats do not directly apply to human responses.

\subsection{Performance Comparison with Previous Benchmarks}
\label{app:comparison_with_prev_benchs}

Table \ref{tab:claude-comparison} compares the performance of Claude Sonnet 3.5, the top‐performing model, on \chartqapro{} against its results on two prior chart‐reasoning benchmarks: ChartQA \cite{masry-etal-2022-chartqa} and CharXiv \cite{wang2024charxivchartinggapsrealistic}.

\begin{table*}[t]
    \centering
    \renewcommand{\arraystretch}{1.2}
    \rowcolors{2}{gray!10}{white}
    \scalebox{0.85}{
    \begin{tabular}{llc}
        \toprule
        \rowcolor{gray!25}
        \textbf{Benchmark} & \textbf{Description} & \textbf{Accuracy (\%)} \\
        \midrule
        \textbf{ChartQA} \cite{masry-etal-2022-chartqa} & Standard benchmark for chart reasoning & 90.50 \\
        \textbf{CharXiv} \cite{wang2024charxivchartinggapsrealistic} & Scientific charts from arXiv, limiting diversity & 60.20 \\
        \chartqapro{} \textcolor{blue}{(Ours)} & Diverse in chart sources, topics, styles, and question types & 55.81 \\
        \bottomrule
    \end{tabular}}
    \caption{
        Performance of Claude Sonnet 3.5 across three chart‐reasoning benchmarks. The lower accuracy on \chartqapro{} (55.81\%) illustrates its increased difficulty compared to ChartQA (90.50\%), highlighting the need for more robust chart understanding capabilities.
    }
    \label{tab:claude-comparison}
\end{table*}

\subsection{Error Analysis}
\label{app:error_analysis}

\begin{figure*}[t]
    \includegraphics[width=\textwidth]{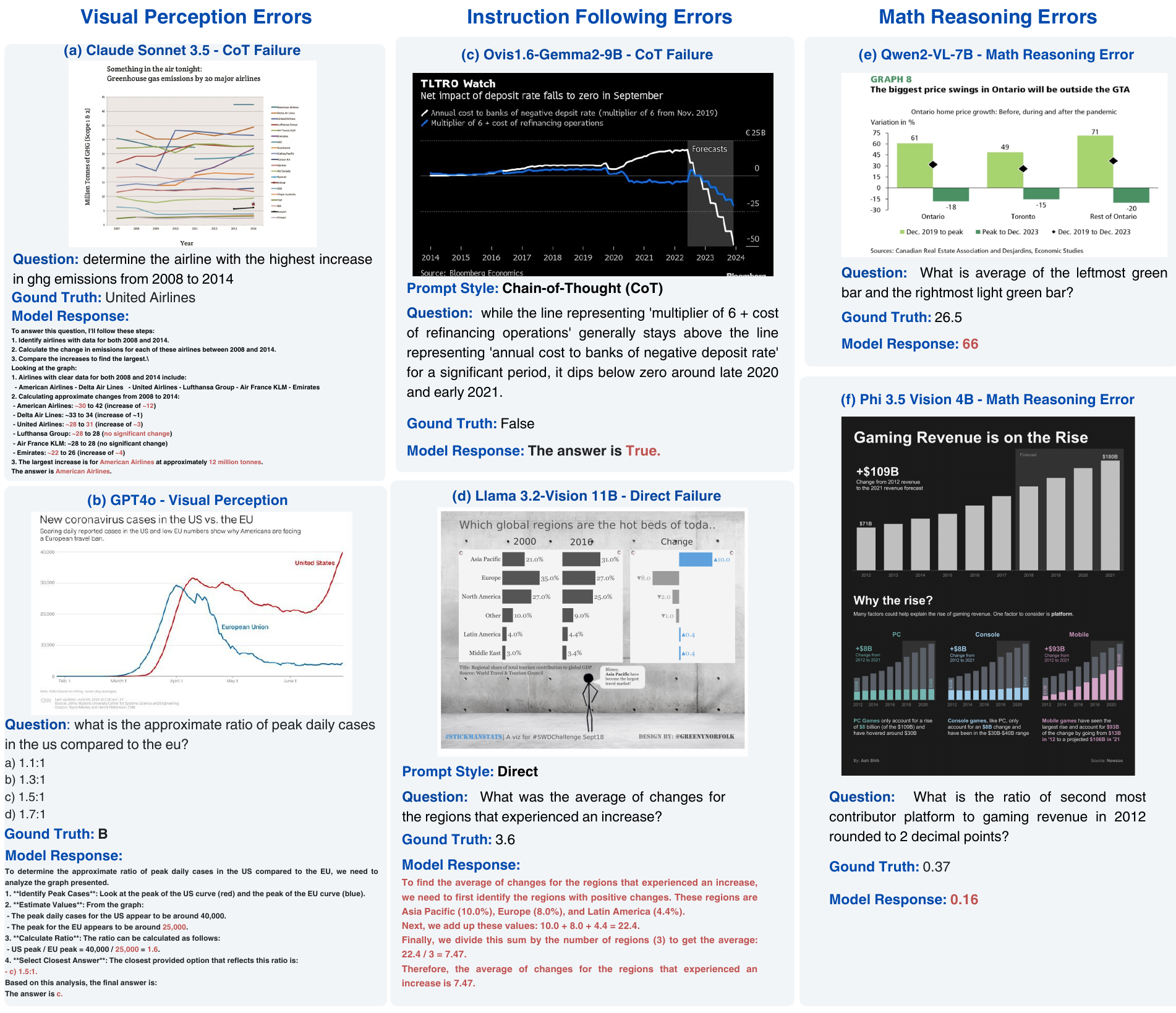}
    \caption{Sample errors across three categories: Visual Perception, Instruction Following (CoT, PoT, Direct), and Mathematical Reasoning.}
    \label{fig:main_error_analysis}
\end{figure*}

Figure \ref{fig:main_error_analysis} presents sample model errors across three categories: visual perception failures, instruction-following issues (CoT, PoT, direct), and mathematical reasoning mistakes.

\subsection{Ablations Results}
\label{app:ablations}

\begin{figure*}[t]
    \includegraphics[width=\textwidth]{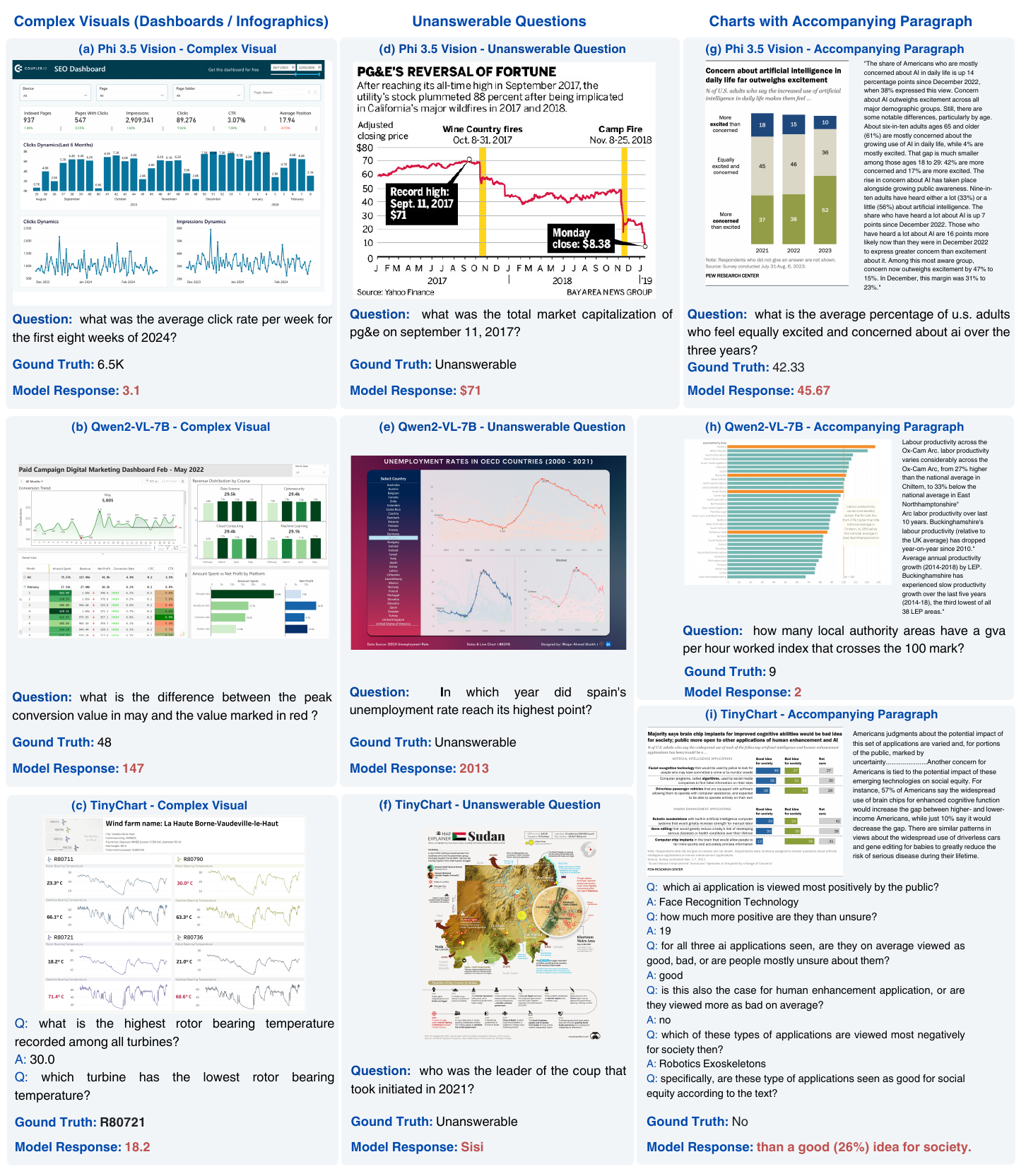}
    \caption{Sample errors from open-source models across different categories in \chartqapro{}.}
    \label{fig:ablations_error_analysis}
\end{figure*}

Figure \ref{fig:ablations_error_analysis} presents sample errors from open-source models—Phi 3.5 Vision 4B \cite{abdin2024phi3technicalreporthighly}, Llama 3.2 Vision 11B \cite{grattafiori2024llama3herdmodels}, and TinyChart \cite{zhang2024tinychartefficientchartunderstanding}—across three categories: complex visuals, unanswerable questions, and charts with accompanying paragraphs.

\end{document}